\documentclass[10pt,twocolumn,letterpaper]{article}

\usepackage{cvpr}              

\usepackage{amsthm,amsmath,amssymb}
\usepackage{graphicx}
\usepackage{mathrsfs}
\usepackage{comment}

\definecolor{cvprblue}{rgb}{0.21,0.49,0.74}
\usepackage[pagebackref,breaklinks,colorlinks,allcolors=cvprblue]{hyperref}
\usepackage{multirow}
\usepackage[normalem]{ulem}
\useunder{\uline}{\ul}{}

\title{Precise, Fast, and Low-cost Concept Erasure in Value Space: \\ Orthogonal Complement Matters }
\author{
    Yuan Wang$^{1}$\footnote[1]\ \ , 
    Ouxiang Li$^1$\footnote[1]\ \ , 
    Tingting Mu$^2$, 
    Yanbin Hao$^3$\footnote[2]\ \ , 
    Kuien Liu$^4$, 
    Xiang Wang$^1$, 
    Xiangnan He$^1$\footnote[2]\ \ \\
    \small $^1$University of Science and Technology of China, 
    $^2$The University of Manchester, \\
    \small $^3$Hefei University of Technology, 
    $^4$ Institute of Software Chinese Academy of Sciences \\
    {\tt\small \{wy1001, lioox\}@mail.ustc.edu.cn, haoyanbin@hotmail.com, xiangnanhe@gmail.com}
}

\begin{document}
\twocolumn[{
	\renewcommand\twocolumn[1][]{#1}
	\maketitle
	\centering
	\vspace*{-0.7cm}
	\includegraphics[width=1\textwidth]{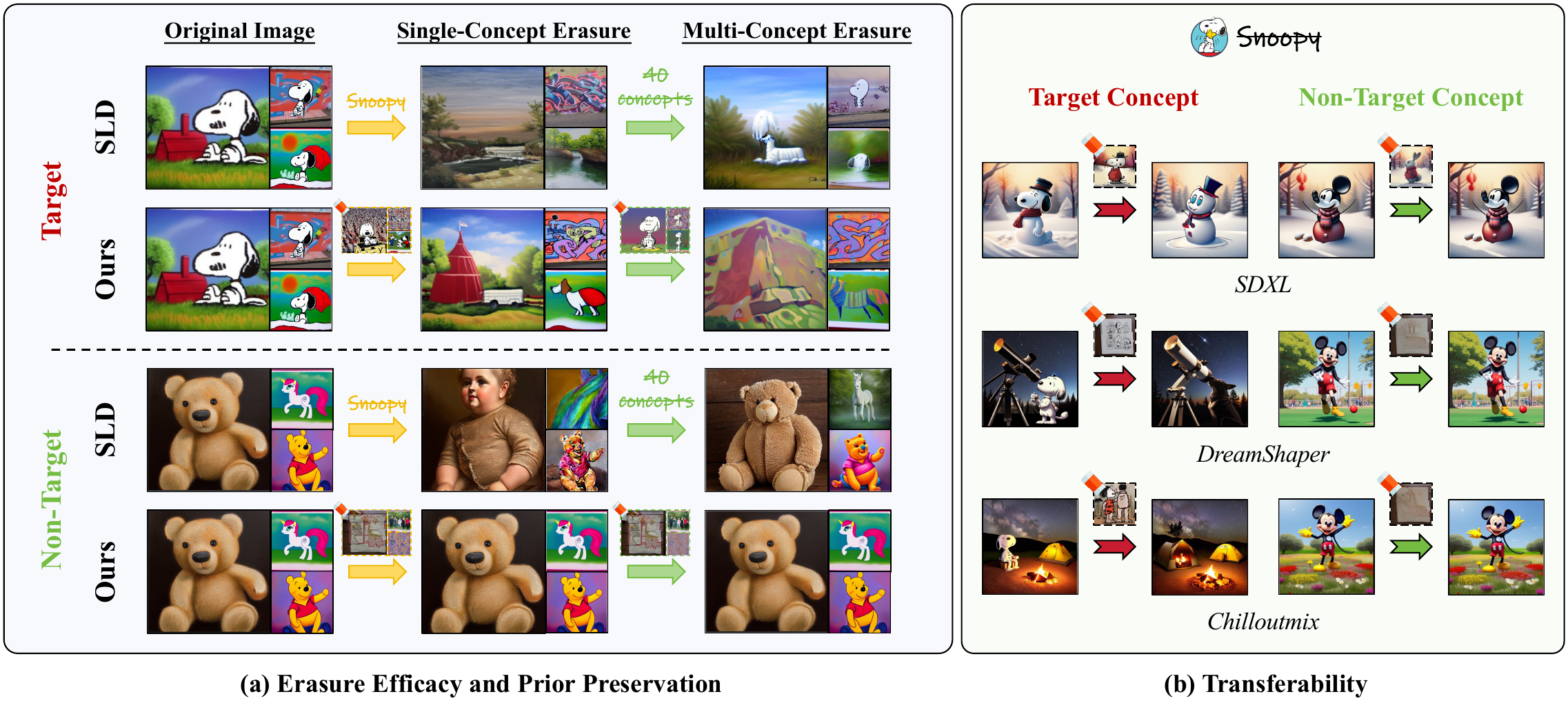} 
	\vspace{-0.7cm}
	\captionof{figure}{The proposed Adaptive Value Decomposer (AdaVD) demonstrates a satisfactory balance between erasure efficacy and prior preservation and an effective transferability across  T2I diffusion models. (a) Compared to SLD \cite{schramowski2023safe}, AdaVD enables precise concept erasure without compromising prior knowledge for non-target concepts at both single- and multi-concept erasure. This is facilitated by a precise disentanglement of target semantics (\eg, \textit{``Snoopy''}) and a robust preservation of non-target ones (\eg, \textit{``Teddy''}), with visualization interpretation marked by \raisebox{-0.6mm}{\includegraphics[height=3.0mm]{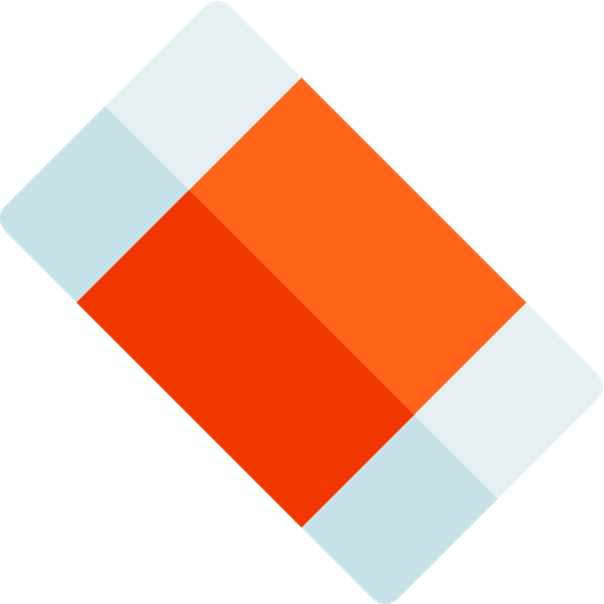}}. (b) AdaVD can be transferred to various T2I models, e.g., SDXL \cite{podellsdxl}, DreamShaper \cite{DreamShaper}, Chilloutmix \cite{Chilloutmix}.}
	\label{fig:intro}
	\vspace*{0.3cm}
}]

\renewcommand{\thefootnote}{\fnsymbol{footnote}} 
\footnotetext[1]{Equal Contributions.}
\footnotetext[2]{Corresponding authors.}
\renewcommand{\thefootnote}{\arabic{footnote}} 
\definecolor{toyred}{RGB}{192, 80, 77}
\definecolor{tableblue}{RGB}{248, 251, 255}
\definecolor{instance_fig_red}{RGB}{252, 232, 234}
\definecolor{instance_fig_blue}{RGB}{218, 227, 245}

\begin{abstract}
Recent success of text-to-image (T2I) generation and its increasing practical applications, enabled by diffusion models, require urgent consideration of erasing unwanted concepts, e.g., copyrighted, offensive, and unsafe ones, from the pre-trained models in a precise, timely, and low-cost manner. 
The twofold demand of concept erasure includes not only a precise removal of the target concept  (i.e., erasure efficacy)  but also a minimal change on non-target content  (i.e., prior preservation), during generation. 
Existing methods face challenges in maintaining an effective balance between erasure efficacy and prior preservation, and they can be computationally costly.
To improve, we propose a precise, fast, and low-cost concept erasure method, called \textbf{Ada}ptive \textbf{V}alue \textbf{D}ecomposer (AdaVD), which is training-free.
Our method is grounded in a classical linear algebraic operation of computing the orthogonal complement, implemented in the value space of each cross-attention layer within the UNet of diffusion models. 
We design a  shift factor to adaptively navigate the erasure strength, 
enhancing effective prior preservation without sacrificing erasure efficacy.
Extensive comparative experiments with both training-based and training-free state-of-the-art methods demonstrate that the proposed AdaVD excels in both single and multiple concept erasure, showing  2  to 10 times improvement in prior preservation than the second best, meanwhile achieving the best or near best erasure efficacy. 
AdaVD supports a series of diffusion models and downstream image generation tasks, with code available on: \href{https://github.com/WYuan1001/AdaVD}{https://github.com/WYuan1001/AdaVD}.
\vspace{-3mm}
\end{abstract}

\vspace{-4mm}
\section{Introduction}
The recent advancements of text-to-image (T2I) diffusion models  \cite{ho2020denoising, nichol2021improved, dhariwal2021diffusion, ho2021classifier, rombach2022high, ye2023ip, zhang2023adding} have enabled users to effortlessly generate high-quality images with simple textual prompts. 
However, such generations would inevitably introduce copyrighted \cite{somepalli2023diffusion, shan2023glaze,carlini2023extracting, li2024improving} or offensive \cite{schramowski2023safe, li2024model} concepts, caused by the noisy training data scraped from web \cite{birhane2021multimodal, schuhmann2021laion}. 
Because it is very costly to re-train large generative models from scratch, it is vital to develop low-cost techniques to precisely erase unwanted semantic concepts in images, \ie, \textit{concept erasure}. 
This task aims at a precise erasure of visual content w.r.t. target concepts from generated images (\ie, \textit{erasure efficacy}), while a faithful preservation of irrelevant content w.r.t. the prompts comprising non-target concepts (\ie, \textit{prior preservation}), safeguarding a secure T2I generation for diffusion models.

A representative category of concept erasure methods is training-based, which fine-tunes a subset of model parameters by formulating meticulous erasing objective functions \cite{kumari2023ablating, gandikota2023erasing, lyu2024one, lu2024mace}. 
Despite good erasure efficacy, they exhibit considerable practical limitations when being deployed. 
For instance, they require expensive individual fine-tuning to erase each concept, which thereby limits their real-time usage.
As an example, it is unacceptable for online T2I platforms to be costly to erase newly emerging concepts, where copyrighted or offensive concepts could arise unexpectedly, with no means to produce a complete list of concepts to erase in advance. 
Moreover, these methods suffer from a limited balance between erasure efficacy and prior preservation, due to their reliance on regularization terms to trade off prior preservation.

An alternative category of concept erasure methods is training-free, such as Negative Prompt (NP) \cite{StableDiffusion}, Safe Latent Diffusion (SLD) \cite{schramowski2023safe}, and SuppressEOT \cite{liget}, which enable real-time erasure. 
They intervene in the image generation process, exhibiting  a range of drawbacks.
For instance, NP was initially designed to enhance image quality and can result in compromised erasure efficacy; while SuppressEOT requires the user to specify the location of the target concept within the prompt, thus, it is not suitable for erasure applications that require full automation. 
Therefore, both NP and SuppressEOT fall short as independent tools for concept erasure. 
Regarding SLD, it does not perform well at retaining prior knowledge of non-target concepts as illustrated in Fig.~\ref{fig:intro}, failing in \textit{precise concept erasure}.
Limited by their drawbacks, current training-free methods are not robust enough to act as an independent concept erasure tool and to be applied in continual   erasure of multiple concepts. 

In this light, we advance concept erasure techniques for T2I generation, by developing a precise, fast, and low-cost method called \textbf{Ada}ptive \textbf{V}alue \textbf{D}ecomposer (AdaVD). 
It is a training-free method, capable of precisely erasing the target concepts and satisfactorily preserving non-target priors with low computational overhead. 
Our core design builds on a classical linear algebraic operation, \ie, projection onto the orthogonal complement of the semantic space of the target concepts.
We conduct this projection-based decomposition in the cross-attention value space, disentangling target semantics from the original prompts.
To improve the erasure precision and prior preservation, we further refine the decomposition by adaptively allocating token-wise shifts.
These shifts are designed to differentiate the strong and specific alignments from the weak and general alignments between prompt tokens and visual content associated with the target concept.
Guided by these shifts, the strong and specific alignments are  erased for precision while the weak and general alignments are retained  for prior preservation.

Fig.~\ref{fig:intro} (a) demonstrates the erased results for both target and non-target concepts. 
It shows that our AdaVD can precisely locate and erase the components indicated by the target semantics, meanwhile keeping the non-target priors maximally unaffected. 
Empirical evaluation shows that AdaVD excels in prior preservation, outperforming the second best by a 2- to 10-fold improvement across various non-target concepts, and meanwhile maintains exceptional erasure efficacy, consistently achieving the best or near-best performance. 
Our contributions are summarized below:
\begin{itemize}
    \item A novel and effective erasing operation by exploiting the projection onto the orthogonal complement of the target concept in the cross-attention value space, to disentangle semantics carried by the target concept.
    \item An adaptive erasing mechanism through a dynamic shift factor, which can effectively minimize the impact on prior knowledge, without compromising the erasure efficacy.
    \item A precise, fast, and low-cost concept erasure technique AdaVD, which works in a training-free manner and supports a series of T2I diffusion models,   capable of  precise concept erasure.
    \item Extensive experiments that demonstrate the superiority of AdaVA against state-of-the-art (SOTA) methods, achieving 2 to 10 times of improvement in prior preservation while maintaining precise erasure efficacy. 
\end{itemize}

\section{Related Works}
\textbf{Re-training and Blocking:} The most straightforward way to erase a target concept from a pre-trained T2I model is to exclude the training data relevant to this concept and re-train the model from scratch, as in Stable Diffusion (SD) v2.0 \cite{StableDiffusion2}.
For this, an Not Safe For Work (NSFW) detector \cite{schramowski2022can, bedapudinudenet, man} can be used to filter unsafe data from LAION-5B \cite{schuhmann2022laion} prior to the training.
However, this approach is time-consuming, requires specialized detectors, and can introduce biases \cite{shi2020improving}.
An  alternative solution is to block the prompts of concerns and restrict the outputs of concerns, by using filters \cite{shi2020improving} and safety checkers \cite{SafetyChecker, rando2022red}. 
Yet, such safeguards are fragile and easy to bypass \cite{yang2024mma, tsairing},  especially when being confronted with crafted malicious prompts \cite{schramowski2023safe}.

\noindent
\textbf{Training-based:} A more effective group of concept erasure solutions   fine-tune   pre-trained generative models, teaching them to  ``forget'' a target concept, e.g.,  Erased Stable Diffusion (ESD) \cite{gandikota2023erasing}.
However, since the ESD design does not consider prior preservation,   both target and non-target concepts are adversely affected. 
To improve, new training techniques have been developed, \eg, ConAbl \cite{kumari2023ablating}, SA \cite{heng2024selective}, UCE \cite{gandikota2024unified}, EraseDiff \cite{wu2024erasediff}, SPM \cite{lyu2024one}  and MACE \cite{lu2024mace}. 
They improve prior preservation through regularization but still fall short in achieving both precise erasure and robust prior preservation. 
Moreover, for every new target concept to be erased, a separate fine-tuning is required.
This is time-consuming, making real-time erasure impractical, especially for highly interactive platforms where unsafe or inappropriate concepts can emerge unexpectedly.

\noindent
\textbf{Training-free:} With largely reduced computing costs, training-free methods are gaining increasing attention. 
NP \cite{StableDiffusion} and SLD \cite{schramowski2023safe} pioneer training-free concept erasure by adjusting classifier-free guidance \cite{ho2021classifier}, leading the generation towards a direction away from the target concepts. 
However, NP lacks fine-grained control over target concepts and compromises prior preservation. 
SLD also compromises the prior knowledge during its generation, disrupting the overall generating quality of non-target concepts.
SuppressEOT \cite{liget}, akin to image editing techniques \cite{hertzprompt}, removes target concepts based on user-specified textual positions.
Its user-involved design makes it more suitable for editing tasks, but not for system-wide erasing tasks that require full automation.
Benefiting from orthogonal decompositions in value spaces, the proposed training-free method AdaVD achieves not only precise concept erasure but also satisfactory prior preservation. 
It supports multi-concept erasure, is compatible across different versions of stable diffusion, and consistently demonstrates superior performance. 
The training-free method  SAFREE \cite{yoonsafree}, concurrent to ours, also uses orthogonal decomposition but operates differently on text embeddings, supported by masking, projection,  Fourier transforms, and a hard control of removal strength.
We compare performance with it in the appendix.

\section{Method}

\begin{figure*}[!t]
\centering
\includegraphics[width=\textwidth]{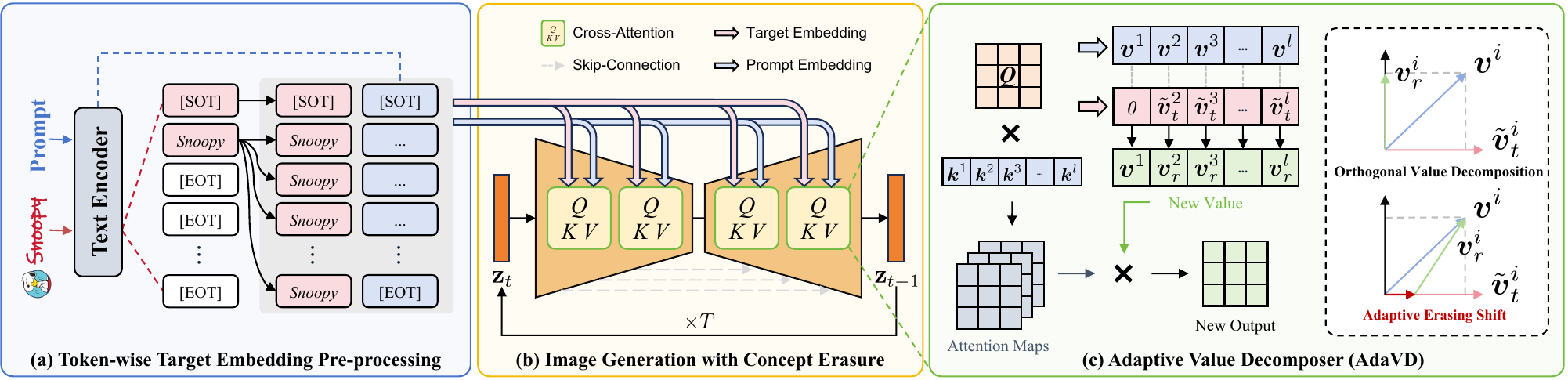}
\vspace{-6mm}
\caption{\textbf{Overview of our Adaptive Value Decomposer (AdaVD)} in erasing the target concept ``\textit{Snoopy}".
(a) First, we token-wisely duplicate the last subject token of the target embedding encoded by the text encoder, except for \texttt{[SOT]}. 
(b) Then, the pre-processed target embedding and corresponding prompt embedding are jointly fed into CA layers within the UNet as conditions, to disentangle target semantics from the original image at each timestep.
(c) In each CA layer, we perform token-wise orthogonal value decomposition with an adaptive token-wise shift. The new value is subsequently multiplied by the attention map, producing the erased output for this CA layer. 
}
\vspace{-3mm}
\label{overview}
\end{figure*}

Concept erasure in T2I generation aims at a successful removal of the visual content indicated by textual concepts (\ie, target concepts), meanwhile a satisfactory preservation of the visual content irrelevant to these concepts (\ie, prior knowledge).
It is challenging to achieve simultaneously satisfactory erasure efficacy and prior preservation.
To advance this field, we propose a precise, fast, and low-cost concept erasure method, termed AdaVD and illustrated in Fig.~\ref{overview} (c).
The method is training-free, and its core design builds on a classical linear algebraic operation, \ie, orthogonal complement.
This simple but elegant geometric operation is able to  guide effectively the image generation, enabling a precise removal of the target concepts from the image content meanwhile a satisfactory preservation of the non-target content.

\subsection{Preliminary on T2I Diffusion Models}
\label{sec:preliminary}
Current T2I models usually include an image compression network \cite{kingma2013auto} and a conditional latent diffusion model \cite{rombach2022high} that performs sequential denoising with a UNet \cite{ronneberger2015u} in the latent space.
The UNet takes as inputs the noise latent variable $\mathbf{z}_t$, timestep $t$, and embedding $\mathbf{C}$ of the textual prompt from a pre-trained CLIP model \cite{radford2021learning}, and predicts the noise $\epsilon_{\theta}(\mathbf{z}_t, t, \mathbf{C})$.
In training-free concept erasure, the noise is additionally conditioned on the target concept with text embedding $\mathbf{C}_{t}$, predicted by $\epsilon_{\theta}(\mathbf{z}_t, t, \mathbf{C}, \mathbf{C}_{t})$. 

Interactions between the image and text modalities are enabled by cross-attention (CA) layers \cite{rombach2022high, vaswani2017attention} within the UNet, which align the latent representation of the noisy image with the semantic detail of the textual prompt. 
Each CA layer computes an attention map $\mathbf{A} = \text{softmax}\left(\frac{\mathbf{Q}\mathbf{K}^{T}}{\sqrt{d}}\right)$ with a latent feature dimension $d$.
The queries $\mathbf{Q}$ are computed from the noisy image features, while both keys $\mathbf{K}$ and values $\mathbf{V}$ from the text embeddings $\mathbf{C}$, using different projection matrices. 
The layer output is a weighted aggregation of $\mathbf{A}$ and $\mathbf{V}$. More details on CA layers are in Appendix \ref{appendix: pre}.

It has been recognized that the keys in CA layers mostly act as the ``Where'' pathway, governing the layout of the attention map and determining the compositional structure of the generated images, while the values as the ``What'' pathway, controlling the content and visual appearance of images \cite{tewel2023key}. 
Because the goal of concept erasure is to modify the visual content of the generated images, we propose to conduct value decompositions into subspaces, which are uniquely constructed by exploiting the target concept and its orthogonal complement in an adaptive fashion. 
We demonstrate that information offered by the orthogonal complement can be used to generate successfully high-quality images with the target concept precisely erased.


\subsection{Token-wise Target Embedding Pre-processing} \label{sec:pre-processing}
Given a textual example, which can be either a target concept to erase or an original prompt, its embedding is computed at the token level by a CLIP text encoder.
Each tokenized example is padded with \texttt{[SOT]} as the prefix and \texttt{[EOT]} at the end, with \texttt{[EOT]} filling any remaining positions to maintain a fixed token length of $l$.
Each token is characterized by an embedding vector of dimension $D_{c}$.

We focus on the embedding of a target concept  $\mathbf{C}_{t} \in \mathbb{R}^{l\times D_c}$, and denote each column of $\mathbf{C}_t^T$  by $ \boldsymbol{c}_t^j$ that  corresponds  to a token vector  where $j\in\{1,2,\ldots, l\}$.
To facilitate a precise erasure, we emphasize the key information carried by a target concept by proposing an embedding duplication.
Specifically, the embedding of the last subject token within its prompt content excluding \texttt{[SOT]} and \texttt{[EOT]}, replaces all the token positions except for \texttt{[SOT]}, as illustrated in Fig.~\ref{overview} (a).  
For instance, the single-token concept ``snoopy'' has a modified embedding matrix corresponding to ``\texttt{[SOT]}, snoopy, snoopy, ..., snoopy'', while the multi-token concept ``Van Gogh'' to  ``\texttt{[SOT]}, gogh, gogh,  ..., gogh''.
Benefiting from the causal attention mechanism used by the CLIP text encoder, the last subject token is able to ``see'' all the prompt content and contains key information \cite{meng2022locating}, therefore is sufficient for erasure calculation.
The modified embedding matrix $\tilde{\mathbf{C}}_t$ is fed into a CA layer to compute the value matrix $\tilde{\mathbf{V}}_t \in \mathbb{R}^{l\times d}$ via a linear projection using the projection matrix $\mathbf{W}_{\textmd{V}}\in \mathbb{R}^{D_{c}\times d}$,   as
\begin{equation}
\label{value_erase}
    \tilde{\mathbf{V}}_t =\tilde{\mathbf{C}}_t\mathbf{W}_{\textmd{V}}= \left[\boldsymbol{c}_t^1,\underbrace{\boldsymbol{c}_t^k, \ldots,    \boldsymbol{c}_t^k}_{l-1} \right]^T\mathbf{W}_{\textmd{V}}.
\end{equation}
The token vector $\boldsymbol{c}_t^1$ corresponds to \texttt{[SOT]} and $\boldsymbol{c}_t^k$ to the last subject token.
The total token number is still $l$.

\subsection{Orthogonal Value Decomposition}
\label{sec_vd}
Given a conditional latent diffusion model trained for standard T2I generation, our proposed erasing operation works by projecting the original textual prompt onto the orthogonal complement of the subspace spanned by the target concepts to erase, and it is implemented in the value space learned at each CA layer of the UNet. 
It supports both single-concept and multi-concept erasure.

%
We do not apply any erasing operation to  \texttt{[SOT]}, as it primarily serves as a prefix and does not carry useful information to distinguish the semantic content.
We start from the modified value matrix $\tilde{\mathbf{V}}_t$ in Eq. (\ref{value_erase}), and use $\tilde{\boldsymbol{v}}_t^j$ to denote the $j$-th column of $\tilde{\mathbf{V}}_t^T$. 
The exclusion of \texttt{[SOT]} in the erasure calculation is equivalent to replacing the value vector in the first row of $\tilde{\mathbf{V}}_t$  by a zero vector, resulting in  $\mathbf{V}_t = \left[\boldsymbol{0},\tilde{\boldsymbol{v}}_t^2, \tilde{\boldsymbol{v}}_t^3, \ldots, \tilde{\boldsymbol{v}}_t^l \right]^T$ to use in the erasing operation.

\subsubsection{Single-concept Erasure}
\label{sec_singlecon}

Our erasing operation works with $\mathbf{V}_t^T$, where each column of $\mathbf{V}_t^T$ corresponds to the erased value vector for token position $j$ and is denoted as $\boldsymbol{v}_t^{j}$, and with the value matrix $\mathbf{V} \in \mathbb{R}^{l\times d}$ computed from the original prompt, where each column of $\mathbf{V}^T$ is referred to as the original value vector for the token position $j$, denoted by $ \boldsymbol{v}^{j}$.
To remove the effect of the target concept from the original prompt, for each token position, we project the original value vector  $ \boldsymbol{v}^{j}$ onto the orthogonal complement of the span of the erased value vector $\boldsymbol{v}_t^{j}$, and denote this orthogonal complement by $\textmd{span}^{\perp}\left(\boldsymbol{v}_t^{j}\right)$. 
This results in the following modified value vector:
\begin{align}
\label{eq:single_res}
    \boldsymbol{v}_r^{j} = \mathbf{P}_{\textmd{span}^{\perp}\left(\boldsymbol{v}_t^{j}\right)}\boldsymbol{v}^{j} 
    =\; & \left(\mathbf{I}_{d} -\mathbf{P}_{\textmd{span}\left(\boldsymbol{v}_t^{j}\right)}\right)\boldsymbol{v}^{j} \\ 
    \nonumber
    = \;& \boldsymbol{v}^{j} - \frac{\boldsymbol{v}_t^{j} \boldsymbol{v}_t^{j}{}^T}{\boldsymbol{v}_t^{j}{}^T\boldsymbol{v}_t^{j}}\boldsymbol{v}^{j} = \boldsymbol{v}^{j} - \frac{\boldsymbol{v}_t^{j}{}^T\boldsymbol{v}^{j}}{\boldsymbol{v}_t^{j}{}^T\boldsymbol{v}_t^{j}}\boldsymbol{v}_t^{j},
\end{align}
where $\mathbf{P}_{\mathbb{X}}\boldsymbol{x}$ denotes the orthogonal projection of a vector $\boldsymbol{x}$ onto the space $\mathbb{X}$, and $\mathbf{I}_{d}$ is an identity matrix of size $d$.
The modified value vector $\boldsymbol{v}_{r}^{j} $ is used, instead of $\boldsymbol{v}^{j}$, to calculate the output of the CA layer, as illustrated in Fig.~\ref{overview} (c).
Since $\boldsymbol{v}_t^{1} = \boldsymbol{0}$,  it has $\boldsymbol{v}_r^{1} = \boldsymbol{v}^{1}$, meaning that no erasure is performed for \texttt{[SOT]}.

\subsubsection{Multi-concept Erasure}
\label{sec_multicon}
We generalize the above operation to erase a set of $n$ target concepts, and their corresponding modified value matrices are denoted by  $\left\{\mathbf{V}_{t}^h \in \mathbb{R}^{l\times d}\right\}_{h=1}^n$.
We use $\boldsymbol{v}_{t}^{h,j}$ to denote the $j$-th column of  $\left(\mathbf{V}_{t}^h\right)^T$, referred to as the erased value vector for the $j$-th token position of the $h$-th target concept.
Our erasing operation can naturally be extended to projecting the original prompt to the orthogonal complement of the span of the $n$ erased value vectors $\left\{\boldsymbol{v}_{t}^{h,j}\right\}_{h=1}^n$, denoted by $\textmd{span}^{\perp}\left(\left\{\boldsymbol{v}_t^{h,j}\right\}_{h=1}^n\right)$.
To calculate the projection, we first conduct the Gram-Schmidt orthogonalization to obtain a set of $n$ orthonormal basis vectors $\left\{\boldsymbol{o}_t^{h,j}\right\}_{h=1}^n$ for the span of $\left\{\boldsymbol{v}_t^{h,j}\right\}_{h=1}^n$. 
Here, we assume the value vectors in $\left\{\boldsymbol{v}_t^{h,j}\right\}_{h=1}^n$ are linearly independent and they form a basis.
Such an assumption is reasonable because the multiple concepts to erase should be semantically different in practice, otherwise, a single-concept erasure would be sufficient. 
The desired projection  is then computed by
\begin{align}
\label{eq:multi_res1}
    \boldsymbol{v}_r^{j} =\; & \mathbf{P}_{\textmd{span}^{\perp}\left(\left\{\boldsymbol{v}_t^{h,j}\right\}_{h=1}^n\right)}\boldsymbol{v}^{j} = \mathbf{P}_{\textmd{span}^{\perp}\left(\left\{\boldsymbol{o}_t^{h,j}\right\}_{h=1}^n\right)}\boldsymbol{v}^{j} \\
    \nonumber
    =\; & \left(\mathbf{I}_d -\mathbf{P}_{\textmd{span}\left(\left\{\boldsymbol{o}_t^{h,j}\right\}_{h=1}^n\right)}\right)\boldsymbol{v}^{j}  \\
     \nonumber
     = \; & \boldsymbol{v}^{j} - \sum_{h=1}^n\left(\boldsymbol{o}_t^{h,j}\right)^T\boldsymbol{v}^{j} \boldsymbol{o}_t^{h,j}.
\end{align}
As an addition, we provide in Appendix \ref{appendix:equal_de} an alternative way to compute $\mathbf{P}_{\textmd{span}^{\perp}\left(\left\{\boldsymbol{v}_t^{h,j}\right\}_{h=1}^n\right)}\boldsymbol{v}^{j}$ that does not require Gram-Schmidt orthogonalization but matrix inverse.

\subsection{Adaptive Erasing Shift}
\label{sec_an}

\begin{figure}[!t]
\centering
\includegraphics[width=\hsize]{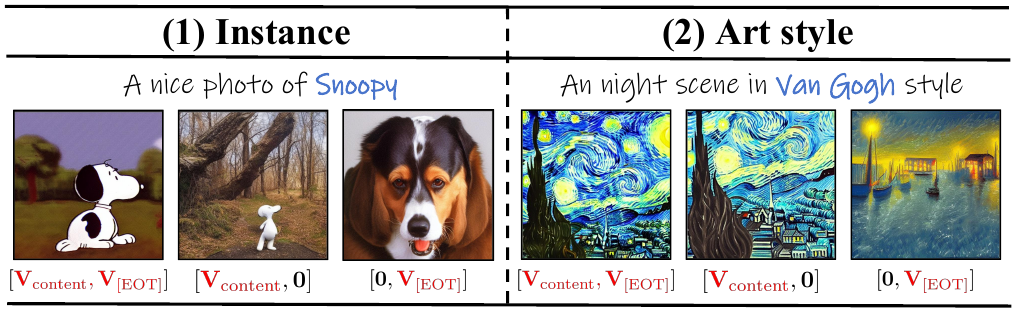}
\vspace{-6mm}
\caption{We analyze the contribution of different tokens in text-visual alignment by separately masking the value of content tokens and \texttt{[EOT]} tokens, where content tokens carry more featured information than those \texttt{[EOT]} tokens.
}
\vspace{-3mm}
\label{fig:toy}
\end{figure}

In practice, given a pair of textual prompts and a target concept to erase, their token-wise relevance can vary across different token positions.
The different tokens of the prompt carry information with different intensities and focuses and thus can have quite different effects on image generation.
As an example, we compare the generated images in each subfigure of Fig.~\ref{fig:toy}, using different versions of the value matrix $\mathbf{V}$ for the same prompt. 
We separate the values corresponding to the \texttt{[EOT]} tokens and the remaining, expressing it as  $\mathbf{V} = [\mathbf{V}_{\textmd{content}}, \mathbf{V}_{\texttt{[EOT]}}]$. Herein, we include   \texttt{[SOT]} within $\mathbf{V}_\textmd{content}$ for simplicity.
We obtain three versions of $\mathbf{V}$ by (1) keeping it as what it is, (2) setting $\mathbf{V}_{\texttt{[EOT]}}$ as zero, and (3) setting $\mathbf{V}_{\textmd{content}}$ as zero.
Fig.~\ref{fig:toy} shows that the prompt content carries more featured information than those \texttt{[EOT]} tokens.
This motivates us to further improve the design in Eqs.~(\ref{eq:single_res}) and (\ref{eq:multi_res1}), by enabling adaptive adjustment of the erasing operation at the token level.

We discover that, although a projection onto the orthogonal complement of the target concept is effective at erasing this concept itself, it can sometimes affect the prior preservation due to an excessive removal of information.
Therefore, our adaptive design is focused on improving the prior preservation.
When semantics carried by a prompt token are less relevant to a target token, we intend to perform less erasure to protect the prior image content. 
According to \cite{qiu2023controlling}, angular information plays a more critical role in conveying semantics than magnitude during image generation.
These motivate us to exploit the cosine similarity between the value vectors of a prompt token and a target token and use it to derive a shift factor to adjust the erasing strength along the identified erasing direction, which we refer to as the erasing shift. 
In general, we let the factor reduce when the cosine similarity becomes smaller.
 
We denote the shift factor by $\delta(\cdot, \cdot): \mathbb{R}^d\times \mathbb{R}^d \rightarrow \mathbb{R}$, which is a function of two value vectors.
Using it, we revise the erasing operation in Eq. (\ref{eq:single_res}) for a single concept to 
\begin{equation}
\label{eq:single_res_scale}
    \boldsymbol{v}_r^{j} =  \boldsymbol{v}^{j} - \frac{\delta\left(\boldsymbol{v}_t^{j}, \boldsymbol{v}^{j} \right)\boldsymbol{v}_t^{j}{}^T\boldsymbol{v}^{j}}{\boldsymbol{v}_t^{j}{}^T\boldsymbol{v}_t^{j}}  \boldsymbol{v}_t^{j}.
\end{equation}
While for multiple concepts, Eq. (\ref{eq:multi_res1}) is revised to
\begin{equation}
\label{eq:multi_res_scale}
    \boldsymbol{v}_r^{j} =  \boldsymbol{v}^{j} - \sum_{h=1}^n \delta\left(\boldsymbol{v}_t^{h,j}, \boldsymbol{v}^{j} \right) \left(\sum_{k=1}^{n}w_{hk}\left(\boldsymbol{o}_t^{k,j}\right)^T\boldsymbol{v}^{j}\right) \boldsymbol{v}_t^{h,j},
\end{equation}
where $w_{hk}$ is the $hk$-th element  of the projection matrix that transforms the basis $\left\{\boldsymbol{v}_t^{h,j}\right\}_{h=1}^n$ to the orthonormal basis $\left\{\boldsymbol{o}_t^{h,j}\right\}_{h=1}^n$. 
For a single target concept, it is straightforward to introduce the shift factor, as in Eq. (\ref{eq:single_res_scale}). 
However, in the case of erasing multiple concepts, the derivation of a shifted operation is less straightforward, as the orthonormal basis does not carry meaningful semantics, thus a transformation back to the value vectors of the target tokens is needed.
We explain in Appendix \ref{appendix:equal_de} how to derive   Eq. (\ref{eq:multi_res_scale}) from Eq. (\ref{eq:multi_res1}).
The revised erasing operations result in the proposal AdaVD.

Our shift factor design builds on the sigmoid function. 
Given two input vectors $\boldsymbol{x}, \boldsymbol{y}\in \mathbb{R}^d$, we formulate it as 
\begin{equation}
\label{eq:shift_factor}
    \delta(\boldsymbol{x}, \boldsymbol{y}) =\frac{s}{1+ e^{-p(\textmd{cos}(\boldsymbol{x}, \boldsymbol{y})-\epsilon)}}.
\end{equation}
The cosine threshold $\epsilon$  allows a strong erasure when exceeding the threshold. 
A negative cosine similarity indicates a very weak relevance between the target and prompt tokens, suggesting no erasure is required.
Therefore, we set  $0<\epsilon<1$ to quantify and filter the relatively weak relevance indicated by a positive but small cosine similarity.
The hyper-parameter $s>0$ controls the factor scale, while $p>0$ controls the increasing rate of the  $\delta$ value against the cosine value.
In Appendix \ref{appendix:hyper}, we provide hyper-parameter implementation details and a comprehensive analysis.
\section{Experiments and Result Analysis}
\label{sec:experiments}
We conduct extensive experiments to evaluate the proposed AdaVD for erasing a diverse range of target concepts, 
covering specific instances, art styles, NSFW content, and celebrity. 
We compare with SOTA training-based methods, including ConAbl \cite{kumari2023ablating}, ESD \cite{gandikota2023erasing}, SPM \cite{lyu2024one}, MACE \cite{lu2024mace}  and SOTA training-free methods including NP \cite{StableDiffusion} and SLD \cite{schramowski2023safe}. 
We also demonstrate the time efficiency and interpretability of AdaVD, along with its wider usage in other downstream image generation tasks, coupled with a series of diffusion models.

\vspace{-1mm}
\subsection{Experimental Setup}
\label{setup}
\textbf{Implementation:} 
We employ SD v1.4 \cite{StableDiffusion} to generate images using the DPM-solver sampler \cite{lu2022dpm} over 30 sampling steps with classifier-free guidance \cite{ho2021classifier} of 7.5. 
All the compared methods are implemented following their default configurations available from their official repository. 
Further implementation details are provided in Appendix \ref{appendix:detailed setting}.

\noindent
\textbf{Evaluation Data:} 
Adopting the same evaluation protocol from SPM \cite{lyu2024one}, we assess the methods based on 80 instance templates, 30 art style templates, and 25 celebrity templates, and benchmark each method by generating 10 images per template per concept in evaluation.
To assess the performance of NSFW erasure, we use the I2P benchmark \cite{schramowski2023safe}.

\noindent
\textbf{Performance Metrics:} We follow the widely used evaluation metrics for concept erasure, including the CLIP score (CS) \cite{radford2021learning} to assess erasure efficacy and the Fréchet inception distance (FID) \cite{heusel2017gans} to assess prior preservation.
 CS  calculates the cosine similarity between a textual prompt and the generated image \cite{radford2021learning}.
We examine two CS values before and after erasing and compare the CS value after the erasure.
When the prompts contain the target concepts, a  more reduced CS indicates a more effective erasure of the target concepts. 
 FID  measures the distance between images generated before and after erasing \cite{heusel2017gans}. 
A lower FID indicates a better alignment between the two images.
Thus, for non-target concepts, lower FID values indicate better prior preservation.
Overall, a precise concept erasure should have a low CS for prompts containing the target concepts and a low FID for prompts composed of non-target concepts.

\noindent
\textbf{Summary:} Results on instance and art style concept erasure are reported in Sections \ref{sec:instance} and \ref{sec:art}, while results on celebrity and NSFW erasure are reported in Appendices \ref{sec:nsfw} and \ref{sec:celebrity}. Additional analyses, including extended quantitative results, comparisons with SuppressEOT \cite{liget}, performance on different versions of SD, and further discussions, are presented in the Appendix.

\vspace{-1mm}
\subsection{On Instance Concept Erasure} 
\label{sec:instance}

\begin{table}[t]
\centering
\renewcommand{\arraystretch}{1.15}  
\resizebox{\hsize}{!}{\begin{tabular}{ccccccc}
\hline
\multicolumn{1}{c|}{Concept}        & \multicolumn{1}{c|}{Snoopy}                                 & \multicolumn{1}{c|}{Mickey}                                 & \multicolumn{1}{c|}{Spongebob}                                & \multicolumn{1}{c|}{Pikachu}                                   & \multicolumn{1}{c|}{Dog}                                & Legislator                             \\ \hline
\multicolumn{1}{c|}{}        & \multicolumn{1}{c|}{CS}                                     & \multicolumn{1}{c|}{CS}                                     & \multicolumn{1}{c|}{CS}                                     & \multicolumn{1}{c|}{CS}                                     & \multicolumn{1}{c|}{CS}                                     & CS                                     \\ \hline
\multicolumn{1}{c|}{SD v1.4} & \multicolumn{1}{c|}{28.51}                                  & \multicolumn{1}{c|}{26.57}                                  & \multicolumn{1}{c|}{27.43}                                  & \multicolumn{1}{c|}{-}                                  & \multicolumn{1}{c|}{-}                                  & -                                  \\ \hline
\multicolumn{7}{c}{\textit{Erase \textbf{Snoopy}}}                                                                                                                                                                                                                                                                                                                                                                                                                       \\ \hline
\multicolumn{1}{c|}{}        & \multicolumn{1}{c|}{\cellcolor[HTML]{F8FBFF}CS ↓}             & \multicolumn{1}{c|}{FID ↓}            & \multicolumn{1}{c|}{FID ↓}            & \multicolumn{1}{c|}{FID ↓}            & \multicolumn{1}{c|}{FID ↓}            & FID ↓           \\ \hline
\multicolumn{1}{c|}{ConAbl}  & \multicolumn{1}{c|}{\cellcolor[HTML]{F8FBFF}25.38}          & \multicolumn{1}{c|}{38.44}          & \multicolumn{1}{c|}{41.59}          & \multicolumn{1}{c|}{29.68}          & \multicolumn{1}{c|}{27.76}          & 27.36          \\
\multicolumn{1}{c|}{MACE}    & \multicolumn{1}{c|}{\cellcolor[HTML]{F8FBFF}\underline{20.78}}          & \multicolumn{1}{c|}{118.01}         & \multicolumn{1}{c|}{111.90}         & \multicolumn{1}{c|}{81.99}          & \multicolumn{1}{c|}{43.27}          & 65.97          \\
\multicolumn{1}{c|}{SPM}     & \multicolumn{1}{c|}{\cellcolor[HTML]{F8FBFF}23.89}          & \multicolumn{1}{c|}{{\ul 33.06}}    & \multicolumn{1}{c|}{{\ul 34.70}}    & \multicolumn{1}{c|}{{\ul 23.89}}    & \multicolumn{1}{c|}{{\ul 19.61}}    & {\ul 18.26}    \\
\multicolumn{1}{c|}{NP}      & \multicolumn{1}{c|}{\cellcolor[HTML]{F8FBFF}23.66}          & \multicolumn{1}{c|}{59.58}          & \multicolumn{1}{c|}{78.74}          & \multicolumn{1}{c|}{52.37}          & \multicolumn{1}{c|}{67.51}          & 55.22          \\
\multicolumn{1}{c|}{SLD}     & \multicolumn{1}{c|}{\cellcolor[HTML]{F8FBFF}27.84}          & \multicolumn{1}{c|}{48.12}          & \multicolumn{1}{c|}{55.36}          & \multicolumn{1}{c|}{38.74}          & \multicolumn{1}{c|}{41.95}          & 49.08          \\
\multicolumn{1}{c|}{Ours}    & \multicolumn{1}{c|}{\cellcolor[HTML]{F8FBFF}{\textbf{20.28}}} & \multicolumn{1}{c|}{\textbf{5.72}}  & \multicolumn{1}{c|}{\textbf{8.56}} & \multicolumn{1}{c|}{\textbf{5.79}} & \multicolumn{1}{c|}{\textbf{2.32}} & \textbf{6.07}  \\ \hline
\multicolumn{7}{c}{\textit{Erase \textbf{Snoopy} and \textbf{Mickey}}}                                                                                                                                                                                                                                                                                                                                                                                                          \\ \hline
\multicolumn{1}{c|}{}        & \multicolumn{1}{c|}{\cellcolor[HTML]{F8FBFF}CS ↓}             & \multicolumn{1}{c|}{\cellcolor[HTML]{F8FBFF}CS ↓}             & \multicolumn{1}{c|}{FID ↓}            & \multicolumn{1}{c|}{FID ↓}            & \multicolumn{1}{c|}{FID ↓}            & FID ↓           \\ \hline
\multicolumn{1}{c|}{ConAbl}  & \multicolumn{1}{c|}{\cellcolor[HTML]{F8FBFF}24.26}          & \multicolumn{1}{c|}{\cellcolor[HTML]{F8FBFF}24.08}          & \multicolumn{1}{c|}{46.32}          & \multicolumn{1}{c|}{39.63}          & \multicolumn{1}{c|}{30.57}          & 27.49          \\
\multicolumn{1}{c|}{MACE}    & \multicolumn{1}{c|}{\cellcolor[HTML]{F8FBFF}{ \underline{20.74}}}    & \multicolumn{1}{c|}{\cellcolor[HTML]{F8FBFF}\underline{20.71}}          & \multicolumn{1}{c|}{51.49}          & \multicolumn{1}{c|}{110.67}         & \multicolumn{1}{c|}{52.07}         & 77.13          \\
\multicolumn{1}{c|}{SPM}     & \multicolumn{1}{c|}{\cellcolor[HTML]{F8FBFF}23.16}          & \multicolumn{1}{c|}{\cellcolor[HTML]{F8FBFF}22.81}          & \multicolumn{1}{c|}{{\ul 41.58}}    & \multicolumn{1}{c|}{{\ul 31.77}}    & \multicolumn{1}{c|}{{\ul 21.96}}    & {\ul 23.69}    \\
\multicolumn{1}{c|}{NP}      & \multicolumn{1}{c|}{\cellcolor[HTML]{F8FBFF}23.59}          & \multicolumn{1}{c|}{\cellcolor[HTML]{F8FBFF}24.85}          & \multicolumn{1}{c|}{81.41}          & \multicolumn{1}{c|}{50.10}          & \multicolumn{1}{c|}{65.93}          & 58.88          \\
\multicolumn{1}{c|}{SLD}     & \multicolumn{1}{c|}{\cellcolor[HTML]{F8FBFF}27.76}          & \multicolumn{1}{c|}{\cellcolor[HTML]{F8FBFF}26.74}          & \multicolumn{1}{c|}{54.59}          & \multicolumn{1}{c|}{39.24}          & \multicolumn{1}{c|}{41.62}          & 50.13          \\
\multicolumn{1}{c|}{Ours}    & \multicolumn{1}{c|}{\cellcolor[HTML]{F8FBFF} \textbf{20.29}}          & \multicolumn{1}{c|}{\cellcolor[HTML]{F8FBFF}{ \textbf{19.93}}}    & \multicolumn{1}{c|}{\textbf{9.34}} & \multicolumn{1}{c|}{\textbf{5.84}} & \multicolumn{1}{c|}{\textbf{2.41}} & \textbf{6.43}  \\ \hline
\multicolumn{7}{c}{\textit{Erase \textbf{Snoopy} and \textbf{Mickey} and \textbf{Spongebob}}}                                                                                                                                                                                                                                                                                                                                                                                          \\ \hline
\multicolumn{1}{c|}{}        & \multicolumn{1}{c|}{\cellcolor[HTML]{F8FBFF}CS ↓}             & \multicolumn{1}{c|}{\cellcolor[HTML]{F8FBFF}CS ↓}             & \multicolumn{1}{c|}{\cellcolor[HTML]{F8FBFF}CS ↓}             & \multicolumn{1}{c|}{FID ↓}            & \multicolumn{1}{c|}{FID ↓}            & FID ↓            \\ \hline
\multicolumn{1}{c|}{ConAbl}  & \multicolumn{1}{c|}{\cellcolor[HTML]{F8FBFF}23.94}          & \multicolumn{1}{c|}{\cellcolor[HTML]{F8FBFF}23.64}          & \multicolumn{1}{c|}{\cellcolor[HTML]{F8FBFF}25.04}          & \multicolumn{1}{c|}{51.20}          & \multicolumn{1}{c|}{31.59}          & 30.03          \\
\multicolumn{1}{c|}{MACE}    & \multicolumn{1}{c|}{\cellcolor[HTML]{F8FBFF}{\underline{20.48}}}    & \multicolumn{1}{c|}{\cellcolor[HTML]{F8FBFF}\underline{20.50}}          & \multicolumn{1}{c|}{\cellcolor[HTML]{F8FBFF}21.59}          & \multicolumn{1}{c|}{99.68}          & \multicolumn{1}{c|}{47.46}          & 70.38          \\
\multicolumn{1}{c|}{SPM}     & \multicolumn{1}{c|}{\cellcolor[HTML]{F8FBFF}22.81}          & \multicolumn{1}{c|}{\cellcolor[HTML]{F8FBFF}22.35}          & \multicolumn{1}{c|}{\cellcolor[HTML]{F8FBFF}\underline{20.82}}          & \multicolumn{1}{c|}{39.83}          & \multicolumn{1}{c|}{{\ul 22.68}}          & {\ul 25.31}    \\
\multicolumn{1}{c|}{NP}      & \multicolumn{1}{c|}{\cellcolor[HTML]{F8FBFF}24.29}          & \multicolumn{1}{c|}{\cellcolor[HTML]{F8FBFF}24.76}          & \multicolumn{1}{c|}{\cellcolor[HTML]{F8FBFF}25.31}          & \multicolumn{1}{c|}{64.75}          & \multicolumn{1}{c|}{65.10}          & 59.33          \\
\multicolumn{1}{c|}{SLD}     & \multicolumn{1}{c|}{\cellcolor[HTML]{F8FBFF}27.84}          & \multicolumn{1}{c|}{\cellcolor[HTML]{F8FBFF}26.71}          & \multicolumn{1}{c|}{\cellcolor[HTML]{F8FBFF}27.60}          & \multicolumn{1}{c|}{{\ul 39.41}}    & \multicolumn{1}{c|}{42.32}    & 49.88          \\
\multicolumn{1}{c|}{Ours}    & \multicolumn{1}{c|}{\cellcolor[HTML]{F8FBFF}\textbf{19.39}}          & \multicolumn{1}{c|}{\cellcolor[HTML]{F8FBFF}{\textbf{19.73}}}    & \multicolumn{1}{c|}{\cellcolor[HTML]{F8FBFF}{\textbf{20.34}}}    & \multicolumn{1}{c|}{\textbf{6.85}} & \multicolumn{1}{c|}{\textbf{2.79}} & \textbf{7.26} \\ \hline
\end{tabular}
}
\vspace{-2mm}
\caption{\textbf{Quantitative comparison of single- and multi-instance erasure.}
The best and second-best results are marked in \textbf{bold} and \underline{underlined}, respectively. Our AdaVD consistently achieves
the lowest CS and the lowest FID in all cases, indicating superior prior preservation without compromising erasure efficacy.}
\vspace{-3mm}
\label{tab:instance}
\end{table}

\begin{figure*}[!t]
\centering
\includegraphics[width=\textwidth]{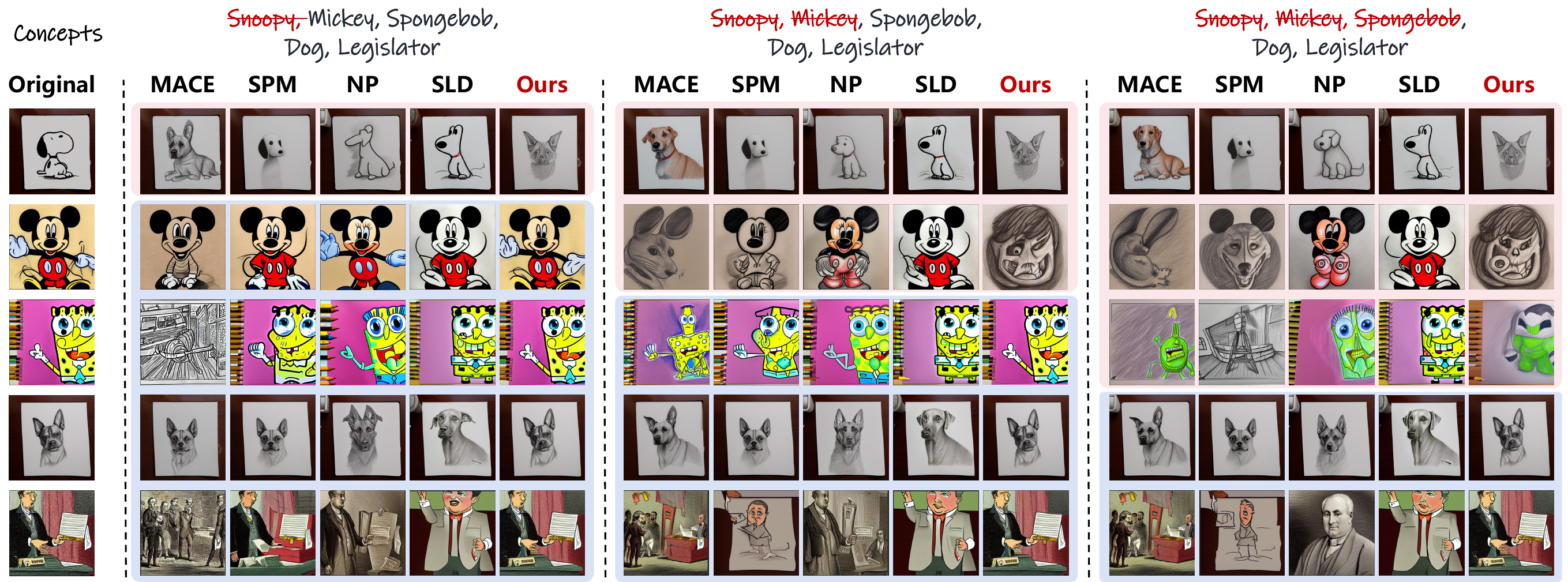}
\vspace{-6mm}
\caption{\textbf{Qualitative comparison of single- and multi-instance erasure.} Both training-based and training-free methods show limitations in prior preservation. In contrast, our AdaVD demonstrates considerable performance in maintaining prior knowledge without compromising erasure efficacy across both single- and multi-concept erasure tasks.}
\vspace{-3mm}
\label{fig:instance}
\end{figure*}

We first conduct the experiment with erasing a single concept \textit{``Snoopy''}.
Six types of prompts were tested, of which one contains \textit{``Snoopy''} and the other five contain only non-target concepts. 
The results are compared in the top block of Table \ref{tab:instance}.
It can be seen that our proposed AdaVD achieves the lowest CS and FID scores in all cases.
Particularly, its FID is   $33\%$ lower than that of the second-best method.
It improves over SOTA by significantly enhanced prior preservation without compromising the erasure precision, as exemplified in Fig.~\ref{fig:instance}.
It can be observed from Fig.~\ref{fig:instance} that methods such as SPM, NP, and SLD fail to fully erase \textit{``Snoopy''} which can be found from the ear and the shape characteristic of the generated image after erasure. 
MACE can successfully erase the  \textit{``Snoopy''} concept.
But all the competing methods suffer from degraded image quality in non-target concept generation, \eg, \textit{``Spongebob''}.

We then compare the performance for multi-concept erasure, experimenting with two cases, of which one erases two concepts of \textit{``Snoopy''} and \textit{``Mickey''} together and the other erases three concepts of \textit{``Snoopy''}, \textit{``Mickey''} and \textit{``Spongebob''} together. 
Results are reported in the bottom two blocks of Table \ref{tab:instance} and visualized in the second and third sections of Fig.~\ref{fig:instance}.
Similarly, our AdaVD achieves the lowest CS and FID scores across all cases.
This verifies that  AdaVD is capable of handling more complex multi-concept erasure and simultaneously fighting against catastrophic forgetting. 
Conversely, other methods struggle to consistently perform well for multi-concept erasure. 
For example, when erasing \textit{``Snoopy''} and \textit{``Mickey''} together, the image quality for non-target concepts like \textit{``Spongebob''} and \textit{``Legislator''} shows apparent degradation in Fig.~\ref{fig:instance}. 
SLD, in particular, fails to maintain its erasure ability even when dealing with 2-concept erasure.
In Appendix \ref{appendix:multi-concept}, we report more experimental results for erasing up to 40 concepts, where our AdaVD can extend to erase dozens of concepts in practice, maintaining consistent erasure efficacy and prior preservation, as shown in Fig.~\ref{fig:intro}.

\vspace{-1mm}
\subsection{On Art Style Erasure} \label{sec:art}
\begin{figure}[!t]
\centering
\includegraphics[width=\hsize]{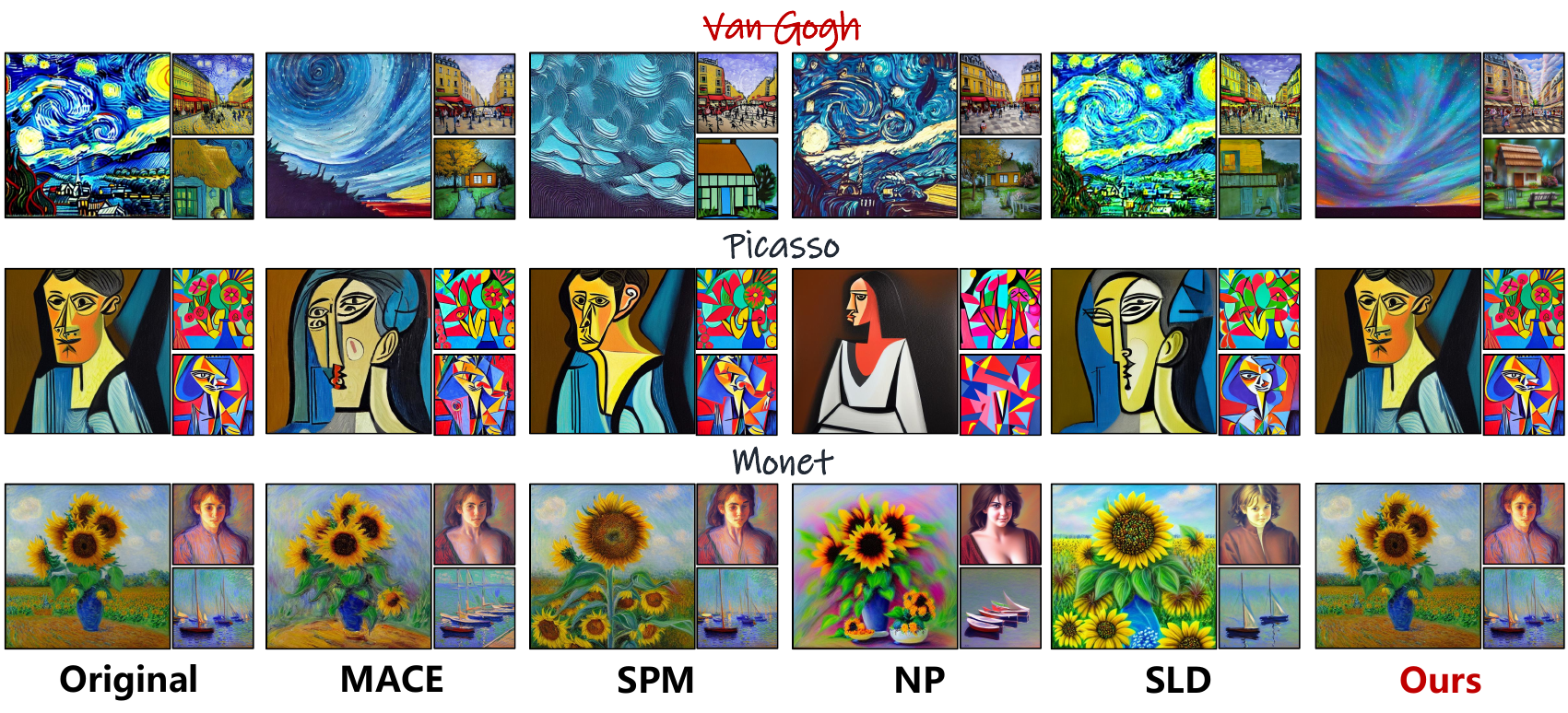}
\vspace{-6mm}
\caption{\textbf{Qualitative comparison of art style erasure.} Our AdaVD can effectively remove the target concept \textit{``Van Gogh"} while preserving non-target styles like \textit{``Picasso"} and \textit{``Monet"}.}
\vspace{-5mm}
\label{fig:art_style}
\end{figure}

\begin{table}[t]
\centering
\renewcommand{\arraystretch}{1.15}  
\resizebox{\hsize}{!}{\begin{tabular}{cccccc}
\hline
\multicolumn{1}{c|}{Concept} & \multicolumn{1}{c|}{Van Gogh}                               & \multicolumn{1}{c|}{Picasso}                                & \multicolumn{1}{c|}{Monet}                                  & \multicolumn{1}{c|}{Andy Warhol}    & Caravaggio    \\ \hline
\multicolumn{1}{l|}{}                   & \multicolumn{1}{c|}{CS}                                     & \multicolumn{1}{c|}{CS}                                     & \multicolumn{1}{c|}{CS}                                     & \multicolumn{1}{c|}{CS}             & CS            \\ \hline
\multicolumn{1}{c|}{SD v1.4}            & \multicolumn{1}{c|}{29.21}                                  & \multicolumn{1}{c|}{29.06}                                  & \multicolumn{1}{c|}{29.02}                                  & \multicolumn{1}{c|}{-}          & -         \\ \hline
\multicolumn{6}{c}{\textit{Erase \textbf{Van Gogh}}}                                                                                                                                                                                                                                                    \\ \hline
\multicolumn{1}{c|}{}                   & \multicolumn{1}{c|}{\cellcolor[HTML]{F8FBFF}CS ↓}             & \multicolumn{1}{c|}{FID ↓}                                    & \multicolumn{1}{c|}{FID ↓}                                    & \multicolumn{1}{c|}{FID ↓}            & FID ↓           \\ \hline
\multicolumn{1}{c|}{ConAbl}             & \multicolumn{1}{c|}{\cellcolor[HTML]{F8FBFF}28.80}          & \multicolumn{1}{c|}{71.71}                                  & \multicolumn{1}{c|}{138.72}                                 & \multicolumn{1}{c|}{70.30}          & 73.10         \\
\multicolumn{1}{c|}{MACE}               & \multicolumn{1}{c|}{\cellcolor[HTML]{F8FBFF}27.74}          & \multicolumn{1}{c|}{65.77}                                  & \multicolumn{1}{c|}{69.79}                                  & \multicolumn{1}{c|}{83.37}          & 75.41         \\
\multicolumn{1}{c|}{SPM}                & \multicolumn{1}{c|}{\cellcolor[HTML]{F8FBFF}\textbf{24.78}} & \multicolumn{1}{c|}{{\ul 62.25}}                            & \multicolumn{1}{c|}{{\ul 32.27}}                            & \multicolumn{1}{c|}{{\ul 58.30}}    & {\ul 61.50}   \\
\multicolumn{1}{c|}{NP}                 & \multicolumn{1}{c|}{\cellcolor[HTML]{F8FBFF}{ 24.90}}    & \multicolumn{1}{c|}{141.56}                                 & \multicolumn{1}{c|}{124.52}                                 & \multicolumn{1}{c|}{127.85}         & 136.32        \\
\multicolumn{1}{c|}{SLD}                & \multicolumn{1}{c|}{\cellcolor[HTML]{F8FBFF}27.48}          & \multicolumn{1}{c|}{103.96}                                 & \multicolumn{1}{c|}{109.11}                                 & \multicolumn{1}{c|}{103.89}         & 119.32        \\
\multicolumn{1}{c|}{Ours}               & \multicolumn{1}{c|}{\cellcolor[HTML]{F8FBFF}\underline{24.87}}          & \multicolumn{1}{c|}{\textbf{6.82}}                          & \multicolumn{1}{c|}{\textbf{2.66}}                          & \multicolumn{1}{c|}{\textbf{8.36}} & \textbf{6.84} \\ \hline
\multicolumn{6}{c}{\textit{Erase \textbf{Picasso}}}                                                                                                                                                                                                                                                     \\ \hline
\multicolumn{1}{c|}{}                   & \multicolumn{1}{c|}{FID ↓}                                    & \multicolumn{1}{c|}{\cellcolor[HTML]{F8FBFF}CS ↓}             & \multicolumn{1}{c|}{FID ↓}                                    & \multicolumn{1}{c|}{FID ↓}            & FID ↓          \\ \hline
\multicolumn{1}{c|}{ConAbl}             & \multicolumn{1}{c|}{58.62}                                  & \multicolumn{1}{c|}{\cellcolor[HTML]{F8FBFF}27.72}          & \multicolumn{1}{c|}{140.34}                                 & \multicolumn{1}{c|}{73.35}          & 67.44         \\
\multicolumn{1}{c|}{MACE}               & \multicolumn{1}{c|}{60.46}                                  & \multicolumn{1}{c|}{\cellcolor[HTML]{F8FBFF}27.11}          & \multicolumn{1}{c|}{49.92}                                  & \multicolumn{1}{c|}{76.10}          & 72.85         \\
\multicolumn{1}{c|}{SPM}                & \multicolumn{1}{c|}{{\ul 38.79}}                            & \multicolumn{1}{c|}{\cellcolor[HTML]{F8FBFF}{\ul 26.69}}    & \multicolumn{1}{c|}{{\ul 7.76}}                             & \multicolumn{1}{c|}{{\ul 52.00}}    & {\ul 51.40}   \\
\multicolumn{1}{c|}{NP}                 & \multicolumn{1}{c|}{111.35}                                 & \multicolumn{1}{c|}{\cellcolor[HTML]{F8FBFF}\textbf{26.14}} & \multicolumn{1}{c|}{91.11}                                  & \multicolumn{1}{c|}{116.24}         & 121.82        \\
\multicolumn{1}{c|}{SLD}                & \multicolumn{1}{c|}{98.21}                                  & \multicolumn{1}{c|}{\cellcolor[HTML]{F8FBFF}27.03}          & \multicolumn{1}{c|}{93.01}                                  & \multicolumn{1}{c|}{97.00}          & 110.05        \\
\multicolumn{1}{c|}{Ours}               & \multicolumn{1}{c|}{\textbf{5.49}}                          & \multicolumn{1}{c|}{\cellcolor[HTML]{F8FBFF}26.99}          & \multicolumn{1}{c|}{\textbf{2.33}}                          & \multicolumn{1}{c|}{\textbf{9.38}} & \textbf{7.05} \\ \hline
\multicolumn{6}{c}{\textit{Erase \textbf{Monet}}}                                                                                                                                                                                                                                                       \\ \hline
\multicolumn{1}{c|}{}                   & \multicolumn{1}{c|}{FID ↓}                                    & \multicolumn{1}{c|}{FID ↓}                                    & \multicolumn{1}{c|}{\cellcolor[HTML]{F8FBFF}CS ↓}             & \multicolumn{1}{c|}{FID ↓}            & FID ↓          \\ \hline
\multicolumn{1}{c|}{ConAbl}             & \multicolumn{1}{c|}{141.52}                                 & \multicolumn{1}{c|}{132.10}                                 & \multicolumn{1}{c|}{\cellcolor[HTML]{F8FBFF}{\ul 24.53}}    & \multicolumn{1}{c|}{208.38}         & 186.26        \\
\multicolumn{1}{c|}{MACE}               & \multicolumn{1}{c|}{76.90}                                  & \multicolumn{1}{c|}{69.35}                                  & \multicolumn{1}{c|}{\cellcolor[HTML]{F8FBFF}26.89}          & \multicolumn{1}{c|}{88.35}          & 81.72         \\
\multicolumn{1}{c|}{SPM}                & \multicolumn{1}{c|}{{\ul 41.03}}                            & \multicolumn{1}{c|}{{\ul 29.71}}                            & \multicolumn{1}{c|}{\cellcolor[HTML]{F8FBFF}27.00}          & \multicolumn{1}{c|}{{\ul 31.90}}    & {\ul 25.99}   \\
\multicolumn{1}{c|}{NP}                 & \multicolumn{1}{c|}{137.21}                                 & \multicolumn{1}{c|}{126.75}                                 & \multicolumn{1}{c|}{\cellcolor[HTML]{F8FBFF}\textbf{24.47}} & \multicolumn{1}{c|}{127.22}         & 135.83        \\
\multicolumn{1}{c|}{SLD}                & \multicolumn{1}{c|}{94.48}                                  & \multicolumn{1}{c|}{92.88}                                  & \multicolumn{1}{c|}{\cellcolor[HTML]{F8FBFF}25.73}          & \multicolumn{1}{c|}{100.90}         & 114.87        \\
\multicolumn{1}{c|}{Ours}               & \multicolumn{1}{c|}{\textbf{6.94}}                         & \multicolumn{1}{c|}{\textbf{6.50}}                         & \multicolumn{1}{c|}{\cellcolor[HTML]{F8FBFF}26.30}          & \multicolumn{1}{c|}{\textbf{8.46}} & \textbf{7.19} \\ \hline
\end{tabular}}
\vspace{-2mm}
\caption{\textbf{Quantitative comparison of art style erasure.} AdaVD achieves a superior balance between erasure efficacy and prior preservation, especially excelling in prior preservation.}
\vspace{-6mm}
\label{tab:style}
\end{table}

We experiment with erasing specific art style, including \textit{``Van Gogh''}, \textit{``Picasso''} and \textit{``Monet''}.
Results are reported in  Table \ref{tab:style}, and visual comparisons are provided in Fig.~\ref{fig:art_style}.
Our AdaVD exhibits superior prior preservation, and achieves the lowest or close-to-lowest CS and FID scores, demonstrating strong prior preservation without sacrificing erasure efficacy.
Other methods show different drawbacks as observed from the results.
For instance, although NP achieves notably better precision in art style removal as compared to instance removal, Fig.~\ref{fig:art_style} shows that it still struggles to fully erase the \textit{``Van Gogh''} style.
Also, SLD fails to erase the \textit{``Van Gogh''} style.
These two methods also degrade generation quality for non-target styles, such as \textit{``Picasso''} and \textit{``Monet''}, showing harmful effects on non-target concepts directly, as evidenced in both Fig.~\ref{fig:art_style} and Table \ref{tab:style}. 
Both MACE and SPM are effective in erasing the target concept, however, their prior preservation is somehow less satisfactory, which is particularly noticeable in the generated images in \textit{``Monet''} style. 
Differently, AdaVD can effectively and consistently remove the target concept and meanwhile preserve satisfactorily the prior content, as confirmed by Table \ref{tab:style} and Fig.~\ref{fig:art_style}.

\begin{table}[t]
\centering
\tiny  
\renewcommand{\arraystretch}{1.15}  
\resizebox{\hsize}{!}{
\begin{tabular}{c|c|c|c|c}
\hline
              & \scalebox{0.9}{\begin{tabular}[c]{@{}c@{}}Data \\ Preparation \end{tabular}} 
              & \scalebox{0.9}{\begin{tabular}[c]{@{}c@{}}Model \\ Finetune \end{tabular}} 
              & \scalebox{0.9}{\begin{tabular}[c]{@{}c@{}}Image \\ Generation \end{tabular}} 
              & \scalebox{0.9}{\begin{tabular}[c]{@{}c@{}}Total Time \end{tabular}} \\ 
\hline
ConAbl        & \scalebox{0.9}{9290}  & \scalebox{0.9}{1120} & \scalebox{0.9}{0.9}  & \scalebox{0.9}{10419} \\
SPM           & \scalebox{0.9}{0}     & \scalebox{0.9}{72850}& \scalebox{0.9}{1.7}  & \scalebox{0.9}{72867} \\
MACE          & \scalebox{0.9}{303}   & \scalebox{0.9}{232}  & \scalebox{0.9}{0.9}  & \scalebox{0.9}{544}   \\
SLD           & \scalebox{0.9}{0}     & \scalebox{0.9}{0}    & \scalebox{0.9}{1.4}  & \scalebox{0.9}{14}    \\ 
\hline
\rowcolor[HTML]{dadada} Ours & \scalebox{0.9}{4} & \scalebox{0.9}{0} & \scalebox{0.9}{1.8} & \scalebox{0.9}{22} \\ 
\hline
\end{tabular}
}
\vspace{-2mm}
\caption{\textbf{Time consumption of 10-concept erasure}. We calculate the time cost (s) to erase 10 concepts and generate 10 images using one NVIDIA A40 GPU. Compared with training-based methods, AdaVD exhibits exceptional efficiency in real-time erasure.}
\vspace{-6mm}
\label{tab:time}
\end{table}

\vspace{-1mm}
\subsection{Further Analysis}
\label{sec:further}

\noindent{\textbf{Time Consumption:}} 
\label{sec:time}
The computational cost of concept erasure primarily arises from three components, including (1)~\textit{data preparation time} required by training-based methods for preparing training data and by AdaVD for basis computation; (2)~\textit{model fine-tuning time} required by training-based methods; and (3)~\textit{image generation time} required by all methods. In Table \ref{tab:time}, we compare the total time consumption of different methods, as well as their time spent on each component.
The two training-free methods of SLD and AdaVD are significantly faster as no fine-tuning is needed.
Our AdaVD costs slightly more time than SLD, \ie, 0.8 extra seconds per image due to its basis computation, and a total of 8 extra seconds for generating 10 images.
But this mild increase yields a significant performance gain, succeeding in precise concept erasure.

\begin{figure}[!t]
\centering
\includegraphics[width=\hsize]{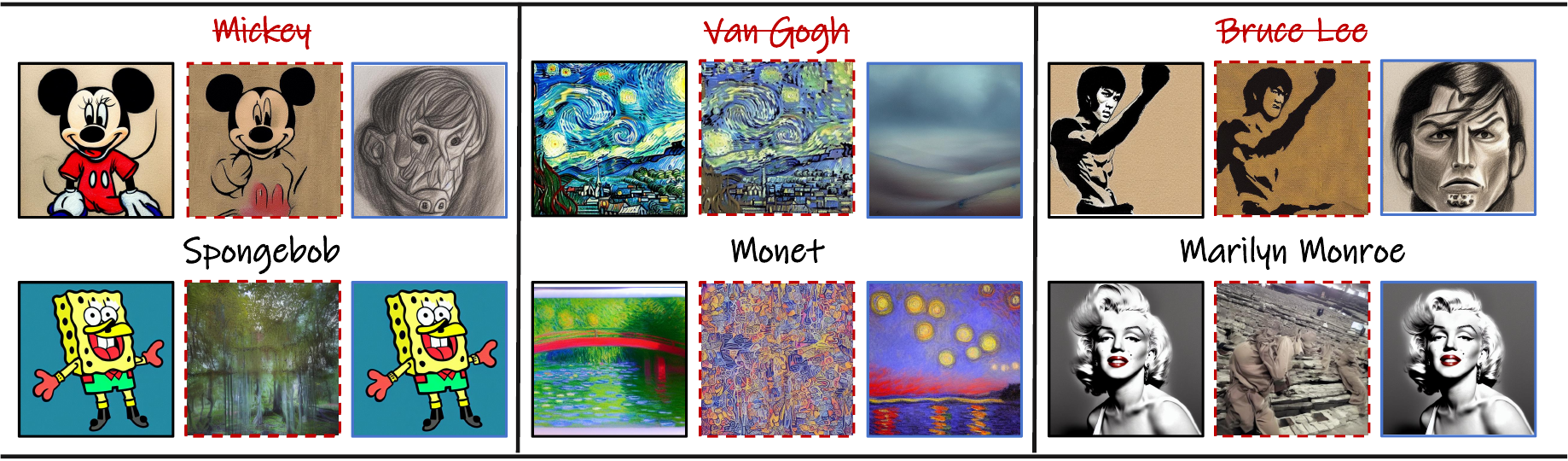}
\vspace{-5mm}
\caption{\textbf{Visualization of erased components.} In each block, we compare both target (1st row) and non-target concept (2nd row) by visualizing the original image (1st column), erased component (2nd column), and generation by our AdaVD (3rd colum).
}
\vspace{-2mm}
\label{fig:interpret}
\end{figure}

\noindent{\textbf{Interpreting  Erased Components by Visualization:}}
\label{sec:inter}
Our AdaVD generates images by replacing the original value vector $\boldsymbol{v}^j$ with $\boldsymbol{v}_r^j$ via orthogonal complement operation. To empirically interpret the rationale behind our method, we visualize the erased component, $\boldsymbol{v}^j - \boldsymbol{v}_r^j$.
Fig.~\ref{fig:interpret} presents three cases of erasing the target concepts  \textit{``Mickey''}, \textit{``Van Gogh''} and \textit{``Bruce Lee''}, where AdaVD successfully erases these target concepts while robustly preserving prior knowledge of non-target concepts.
In the first row, as shown in the middle column of each block, the erased components consistently align with the corresponding target semantics when dealing with the target concepts.
In the second row of non-target concepts, conversely, the erased components do not contain any informative pattern, indicating that they carry no meaningful semantics, and therefore exert minimal impact on the prior knowledge.

\begin{figure}[!t]
\centering
\includegraphics[width=\hsize]{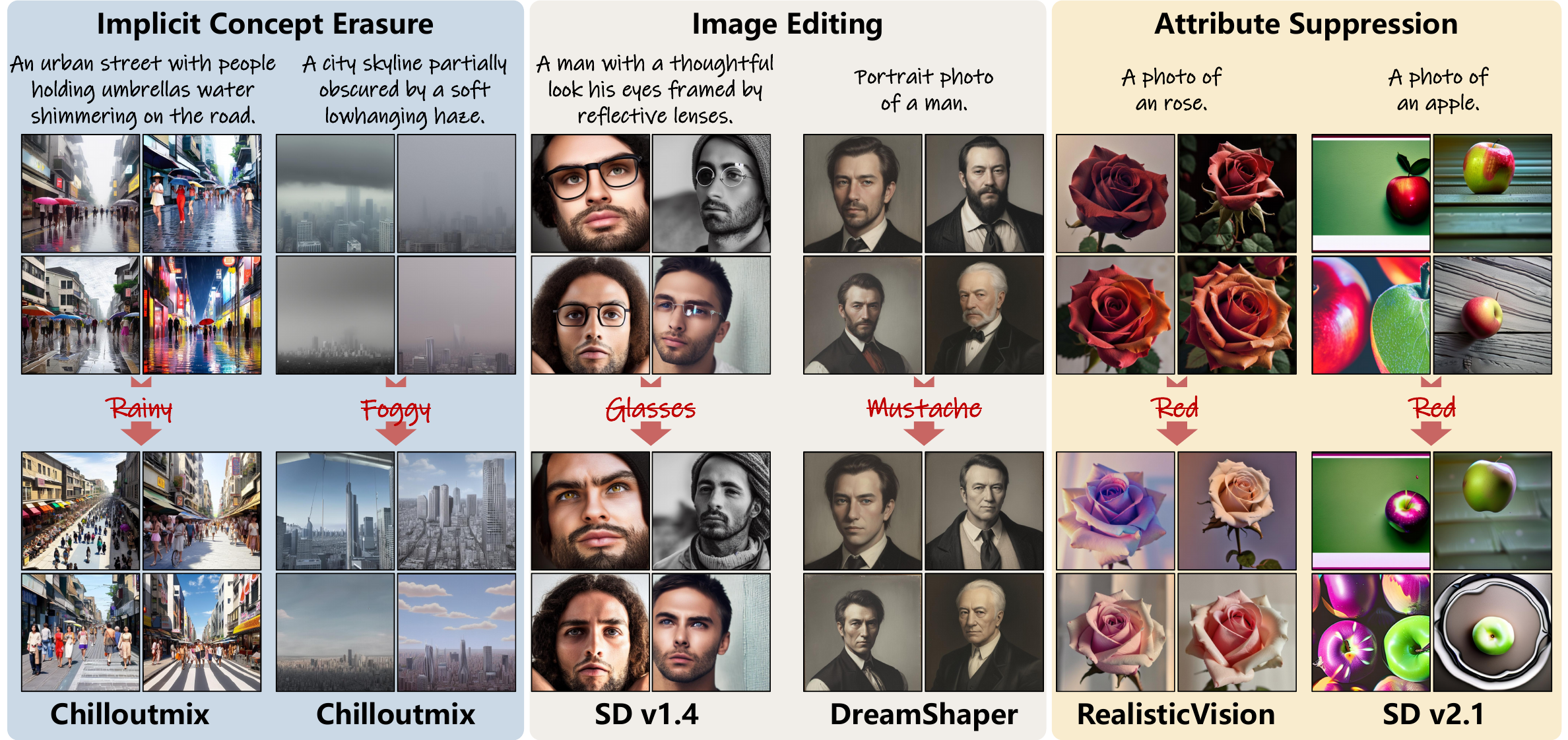}
\vspace{-5mm}
\caption{\textbf{Downstream applications.} We extend AdaVD to versatile generation tasks, including (1) implicit concept erasure, (2) image editing, and (3) attribute suppression, indicating its significant potential for broad applications.
}
\vspace{-6mm}
\label{fig:further}
\end{figure}

\noindent{\textbf{Downstream Applications:}} 
We conduct additional experiments to showcase the versatile applications of our AdaVD across various generative tasks, including (1)~implicit concept erasure: by removing the implicit concepts of \textit{``rainy''} and \textit{``foggy''}; (2)~image editing: by removing the appearance concepts of \textit{``glasses''} and \textit{``mustache''}; and (3)~attribute suppression: by removing the coupled color concept of \textit{``red''}.
Additionally, we integrate AdaVD with a series of diffusion models, including Chilloutmix \cite{Chilloutmix}, DreamShaper \cite{DreamShaper}, RealisticVision \cite{RealisticVsion}, and SD v2.1 \cite{StableDiffusion2.1}, alongside SD v1.4.
As illustrated in Fig.~\ref{fig:further}, despite the absence of explicit mention for \textit{``rainy''} and \textit{``foggy''}, AdaVD can still effectively erase these concepts in image semantic space. 
Meanwhile, AdaVD also precisely removes \textit{``glasses''} and \textit{``mustache''} with minimal changes to other details, highlighting its potential in image editing applications.
For attribute suppression, AdaVD successfully eliminates the color attribute \textit{``red''} from objects such as apples and roses, demonstrating its capability to decouple strongly coupled concepts, \eg, ``roses are red'' embedded in the model's prior knowledge.

\label{sec:further_application}
\section{Conclusion and Future Work}
We have presented AdaVD, a precise, fast, and low-cost method for erasing unwanted concepts.
The idea of leveraging the classical linear algebraic orthogonal complement operation and an adaptive erasing shift design is novel, and has successfully achieved a precise concept erasure. 
Extensive experiments have demonstrated both high erasure efficacy and strong prior preservation of AdaVD for both single- and multi-concept erasure. 
Moreover, AdaVD exhibits excellent interpretability through visualizing its erased components and strong capability in solving downstream tasks.
It has been an intriguing discovery that the orthogonal complement of the value vectors of the target concepts is effective at erasing their inherent semantics.
Despite the empirical success, we will seek to establish a rigorous theoretical understanding of the accumulative effect of applying this linear algebraic operation layer-wise for concept erasure in the future.

\section*{Acknowledgements}
This work is supported by the National Science and Technology Major Project (2023ZD0121102) and the National Natural Science Foundation of China (U24B20180)

{
    \small
    \bibliographystyle{ieeenat_fullname}
    \bibliography{main}
}

\appendix
\onecolumn

\section{Extra Preliminary on CA Layers}
\label{appendix: pre}
The cross-attention (CA) layers in the conditional denoising UNet $\epsilon_{\theta}(\mathbf{z}_t, t, \mathbf{C}) $ align the latent representation of the noisy image with that of the textual prompt. 
The latent variable at time step $t$ is denoted by $\mathbf{z}_t \in \mathbb{R}^{D_{z} \times H \times W} $ with a spatial dimension $H \times W$ and a channel dimension $D{z}$, while the text embedding, i.e. the latent representation of the textual prompt,  is denoted by $\mathbf{C}  \in   \mathbb{R}^{l\times D_c}$.
At the $i$-th CA layer, the attention map is computed by 
\begin{equation}
    \mathbf{A}_{i} = \text{softmax}\left(\frac{\mathbf{Q}_{i} \mathbf{K}_{i}^\top}{\sqrt{d_{i}}}\right),
\end{equation}
where $d_{i}$ is the latent feature dimension.
The queries $\mathbf{Q}_{i} \in \mathbb{R}^{H_{i}W_{i}\times d_{i}}$ are obtained by projecting the latent features of the noisy image returned by the previous module, while both the keys $\mathbf{K}_{i}\in \mathbb{R}^{l\times d_{i}}$ and values $\mathbf{V}_{i} \in \mathbb{R}^{l\times d_{i}}$ are computed by projecting the text embedding but using different projection matrices.
Finally, the output of this CA layer is computed from the attention map, and the values by  $\mathbf{z}_t^{i+1} = \phi\left(\mathbf{A}_{i} \mathbf{V}_{i}\right)$, where a common choice of $\phi(\cdot)$ is a multi-layer perceptron.  
The subsequent modules take $\mathbf{z}_t^{i+1}$ for further processing. For the convenience of explaining, we do not distinguish the notation $i$ between layers in the main text.

\section{Equation Derivation}
\label{appendix:equal_de}
\subsection{On Equation (\ref{eq:multi_res_scale})}

Working with the subspace constructed as the span of the vector set $\left\{\boldsymbol{v}_t^{h,j}\right\}_{h=1}^n$,   we obtain a set of orthonormal basis $\left\{\boldsymbol{o}_t^{h,j}\right\}_{h=1}^n$ through the Gram-Schmidt orthogonalization.
When the value vectors $\left\{\boldsymbol{v}_t^{h,j}\right\}_{h=1}^n$ are linearly independent, each orthonormal basis can be expressed as  a linear combination of these vectors   such that 
\begin{equation}
\label{eq:linear}
    \boldsymbol{o}_t^{h,j} = \sum_{k=1}^n w_{hk}\boldsymbol{v}_t^{k,j},
\end{equation}
where $w_{hk}$ are the combination weights. 
We have explained the linear  independence assumption on $\left\{\boldsymbol{v}_t^{h,j}\right\}_{h=1}^n$ in Section \ref{sec_multicon}.
Incorporating Eq. (\ref{eq:linear}) into Eq. (\ref{eq:multi_res1}) but replacing only the second $\boldsymbol{o}_t^{h,j}$, it results in the following revised calculation of the orthogonal complement: 
\begin{equation}
    \boldsymbol{v}_r^{j} =  \boldsymbol{v}^{j} - \sum_{h=1}^n \left(\sum_{k=1}^{n}w_{hk}\left(\boldsymbol{o}_t^{k,j}\right)^T\boldsymbol{v}^{j}\right) \boldsymbol{v}_t^{h,j}.
\end{equation}
The importance of this revised equation lies in the fact that it computes a weighted sum of the value vectors when performing the erasing.
This enables the application of the adaptive erasing shift mechanism based on the value vectors, for which we further revise the erasing operation as  
\begin{equation}
    \boldsymbol{v}_r^{j} =  \boldsymbol{v}^{j} - \sum_{h=1}^n \delta\left(\boldsymbol{v}_t^{h,j}, \boldsymbol{v}^{j} \right) \left(\sum_{k=1}^{n}w_{hk}\left(\boldsymbol{o}_t^{k,j}\right)^T\boldsymbol{v}^{j}\right) \boldsymbol{v}_t^{h,j}.
\end{equation}
Storing  the combination weights in the matrix $\mathbf{W}=[w_{hk}]\in R^{n\times n}$, it acts as   a  projection matrix transforming the two vector sets by
\begin{equation}
\begin{bmatrix}
   \boldsymbol{o}_t^{1,j} & \cdots & \boldsymbol{o}_t^{n,j} 
\end{bmatrix} 
=
\begin{bmatrix}
   \boldsymbol{v}_t^{1,j} & \cdots & \boldsymbol{v}_t^{n,j}
\end{bmatrix}  \mathbf{W}.
\end{equation}

\subsection{Alternative Orthonormal Basis Calculation}

Purely for the interest of readers, we point out an alternative way to calculate the orthonormal basis.
Constructing a matrix $\hat{\mathbf{V}}_t^j \in \mathbb{R}^{d \times n}$ by using $\left\{\boldsymbol{v}_{t}^{h,j}\right\}_{h=1}^n$ as its columns, following Equation 
(5.13.6) of the linear algebra textbook \cite{meyer2023matrix}, the projection of $\boldsymbol{v}^{j}$ onto $\textmd{span}^{\perp}\left(\left\{\boldsymbol{v}_t^{h,j}\right\}_{h=1}^n\right)$ can be directly computed from $\hat{\mathbf{V}}_t^j$ by 
\begin{align}
\label{eq:multi_res2}
    \boldsymbol{v}_r^{j} =\; & \mathbf{P}_{\textmd{span}^{\perp}\left(\left\{\boldsymbol{v}_t^{h,j}\right\}_{h=1}^n\right)}  \boldsymbol{v}^{j} \\
    \nonumber
    =\; & \left(\mathbf{I}_{d} - \hat{\mathbf{V}}_t^j \left(\left(\hat{\mathbf{V}}_t^j\right)^{T}\hat{\mathbf{V}}_t^j\right)^{-1}\left(\hat{\mathbf{V}}_t^j\right)^{T}\right)\boldsymbol{v}^{j}. 
\end{align}
Compared to Eq. (\ref{eq:multi_res1}),  Eq. (\ref{eq:multi_res2}) does not require the Gram-Schmidt orthogonalization, but the inverse calculation. 
Defining $\mathbf{P}_t^j = \hat{\mathbf{V}}_t^j \left(\left(\hat{\mathbf{V}}_t^j\right)^{T}\hat{\mathbf{V}}_t^j\right)^{-1}\left(\hat{\mathbf{V}}_t^j\right)^{T}$, one potential way to enable token-wise adaptive erasing shift based on Eq. (\ref{eq:multi_res2}) is 
\begin{equation}
    \boldsymbol{v}_r^{j} = \left(\mathbf{I}_{d} 
    - \text{Diag}\left[\delta\left(\boldsymbol{v}_t^{h,j}, \boldsymbol{v}^{j} \right)\right] \mathbf{P}_t^j\right)\boldsymbol{v}^{j},
\end{equation}
where $\text{Diag}\left[\delta\left(\boldsymbol{v}_t^{h,j}, \boldsymbol{v}^{j} \right)\right]$ is a diagonal matrix with shift factors $\left[\delta\left(\boldsymbol{v}_t^{h,j}, \boldsymbol{v}^{j} \right)\right]$ as its diagonal elements.
We leave the in-depth investigation of exploiting this operation in practice to our future work.

\section{Additional Experimental Details}\label{sec:details}
\subsection{On Implementation}
\label{appendix:detailed setting}
To implement SD v1.4,  the DPM-solver is chosen as the sampler, with a total of 30 sampling timesteps and a classifier-free guidance scale set of 7.5. Notably, we set the unconditional prompt to null text, as the negative prompt serves as a training-free method that can be directly compared with our AdaVD. 
To ensure a fair comparison, particularly for prior preservation, we use the same random seed (seed 0) across all methods to generate images under identical conditions. 
For the specific instance, art style, and celebrity erasure, we simply fix the hyperparameters to $p=100$, $\epsilon=0.93$, and $~s=2$.
Our AdaVD performs consistently well with this unified hyper-parameter configuration.

\begin{figure*}[!t]
\centering
\includegraphics[width=\textwidth]{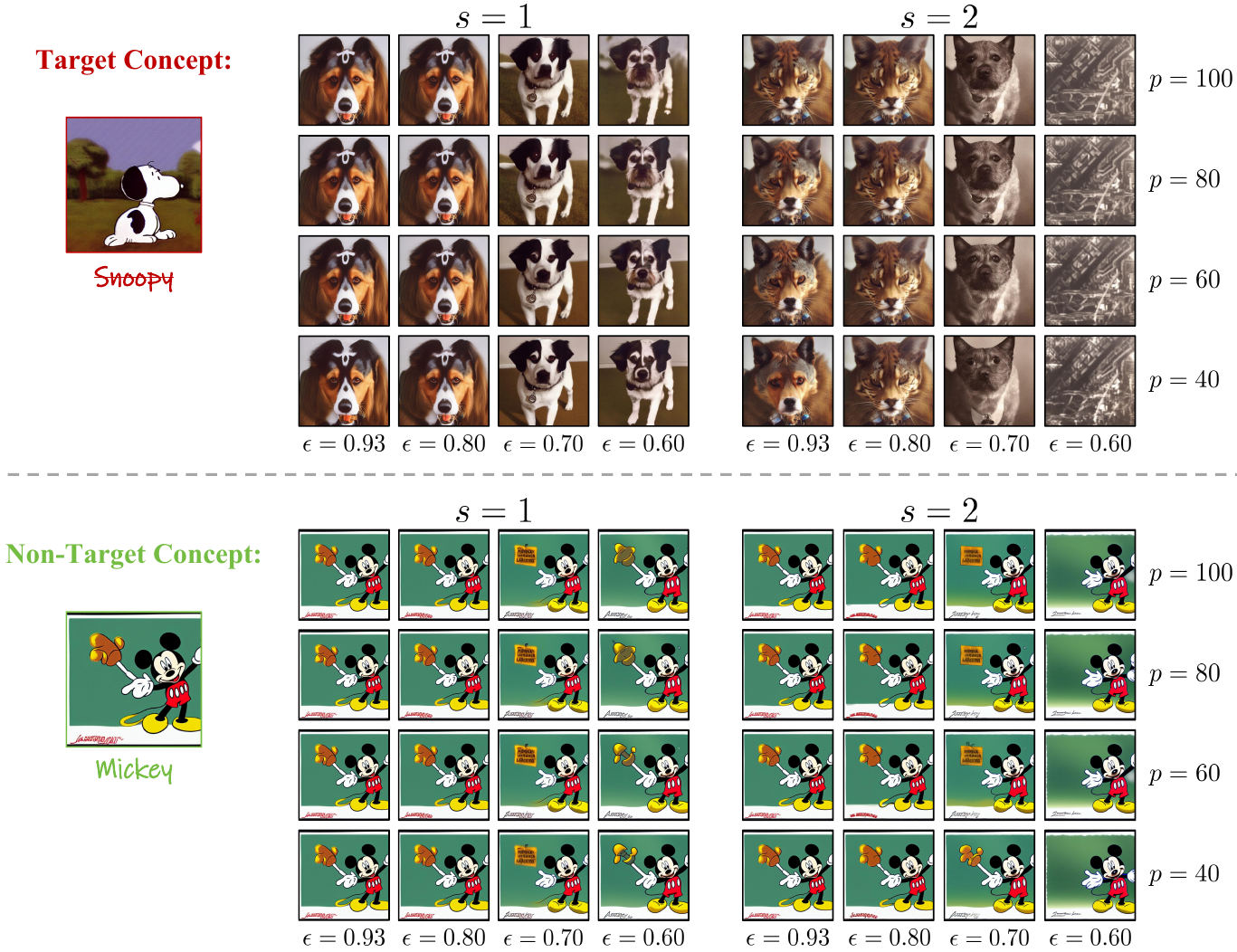}
\caption{\textbf{Impact of hyperparameter settings on erasure efficacy and prior preservation.} To evaluate how hyperparameters affect this balance, we visualize images generated by AdaVD under various hyperparameter settings for the target concept \textit{``Snoopy”} and the related but non-target concept \textit{``Mickey”}.}
\label{fig:param}
\end{figure*}

\subsection{Additional Hyper-parameter Analysis}
\label{appendix:hyper}
The hyper-parameters of the shift factor, including $0<\epsilon<1$ and $p,s>0$, are closely related to the cosine similarities between tokens of the target concepts and tokens of the prompt.
When erasing instances, art styles, and celebrity concepts, we notice that certain non-target concepts contained by the prompt semantically correlate with the target concept, with fairly strong correlations. 
For example, the non-target concept \textit{``Mickey”} exhibits a relatively large cosine similarity of 0.65 with the target concept \textit{``Snoopy”}, as they both belong to the category of cartoon characters. 
This makes it a fine balance between an unaffected generation of these non-target concepts and a successful erasure of the target concept. 
To examine how the erasure strength impacts such a balance, we show in  Fig. \ref{fig:param} different image examples generated by AdaVD under various hyperparameter settings,  for the target concept \textit{``Snoopy”} and non-target concept \textit{``Mickey”}.

Overall, the factor scale $s$ and the threshold $\epsilon$ significantly impact the balance between the erasure efficacy and prior preservation. 
Specifically, it can be observed, from the top part of Fig. \ref{fig:param} (on target concept), that a reduction in $\epsilon$ results in a greater deviation in the generated images as compared to the original,  for content relevant to the target concept.
This indicates an enhanced erasure efficacy. 
This effect is further amplified as $s$ increases. When adopting the setting of $s=2$ and $\epsilon=0.6$, the erasure becomes excessive. 
Conversely, for non-target concept generation, a lower threshold $\epsilon$ can negatively impact the non-target prior, as observed from the bottom part of Fig. \ref{fig:param} (on the non-target concept).
Such a negative impact on non-target concept generation intensifies with increasing $s$, since a larger $s$ amplifies the token shift.
%
This results in a larger divergence from the original token direction, and eventually more noticeable changes in the generated images.

The erasure performance is less sensitive to $p$, but it still has some mild impact. 
For instance, when using a higher value of $s$, a lower $p$ can mitigate changes in the generated visual content that is relevant to the non-target concepts. 
This is demonstrated in the bottom-right part of Fig. \ref{fig:param}. 
When $\epsilon$ decreases to 0.7, setting $p$ to 40 results in less deviation from the original images as compared to other values.
On the other hand, when $s=1$, a higher $p$ positively affects the preservation of some non-target concepts that are related. This is shown in the bottom-left part of Fig. \ref{fig:param}. When $\epsilon=0.6$, the deviation from the original image decreases as $p$ increases from 40 to 100.

\section{Additional  Single-concept Experiments} 
\label{sec:additional}
\subsection{Extended Quantitative Results on Instance and Art Style Erasure}

We present the extended quantitative results on instance erasure and art style erasure in Table \ref{tab:instance_extend} and \ref{tab:style_extend}, respectively. In addition to the CS for the target concept and FID for non-target concepts, we also include the FID for the target concept and CS for non-target concepts. Specifically, FID measures the distribution distance of generated images aligned with the target concept before and after concept erasure, while CS evaluates the semantic consistency between the text prompt of the non-target concept and the generated image after erasure.

However, a lower FID for the target concept only indicates significant visual changes in the generated images but does not confirm that the semantics aligned with the target concept have been fully eliminated. Similarly, a higher CS for the non-target concept suggests that the generated image after concept erasure still aligns closely with the text prompt, but does not guarantee small pixel-level changes. In summary, FID for the target concept and CS for the non-target concept cannot directly measure the effectiveness of erasure or prior preservation. Nevertheless, they remain valuable for further verifying and comparing the erasure efficacy and prior preservation.

\begin{table*}[t]
\centering
\renewcommand{\arraystretch}{1.20}  
\resizebox{0.70\hsize}{!}{\begin{tabular}{ccccccccccccc}
\hline
\multicolumn{1}{c|}{}        & \multicolumn{2}{c|}{Snoopy}                                                                 & \multicolumn{2}{c|}{Mickey}                                                                 & \multicolumn{2}{c|}{Spongebob}                                                              & \multicolumn{2}{c|}{Pikachu}                                      & \multicolumn{2}{c|}{Dog}                                          & \multicolumn{2}{c}{Legislator}               \\ \hline
\multicolumn{1}{c|}{}        & CS                                     & \multicolumn{1}{c|}{FID}                           & CS                                     & \multicolumn{1}{c|}{FID}                           & CS                                     & \multicolumn{1}{c|}{FID}                           & CS                           & \multicolumn{1}{c|}{FID}           & CS                           & \multicolumn{1}{c|}{FID}           & CS                           & FID           \\ \hline
\multicolumn{1}{c|}{SD v1.4} & 28.49                                  & \multicolumn{1}{c|}{-}                             & 26.50                                  & \multicolumn{1}{c|}{-}                             & 27.30                                  & \multicolumn{1}{c|}{-}                             & 27.41                        & \multicolumn{1}{c|}{-}             & 24.27                        & \multicolumn{1}{c|}{-}             & 23.73                        & -             \\ \hline
                             & \multicolumn{12}{c}{\textit{Erase \textbf{Snoopy}}}                                                                                                                                                                                                                                                                                                                                                                                                                                            \\ \hline
\multicolumn{1}{c|}{}        & \cellcolor[HTML]{F8FBFF}CS $\downarrow$            & \multicolumn{1}{c|}{{\color[HTML]{939393} FID}}    & {\color[HTML]{939393} CS}              & \multicolumn{1}{c|}{FID $\downarrow$}                           & {\color[HTML]{939393} CS}              & \multicolumn{1}{c|}{FID $\downarrow$}                           & {\color[HTML]{939393} CS}    & \multicolumn{1}{c|}{FID $\downarrow$}           & {\color[HTML]{939393} CS}    & \multicolumn{1}{c|}{FID $\downarrow$}           & {\color[HTML]{939393} CS}    & FID  $\downarrow$         \\ \hline
\multicolumn{1}{c|}{ConAbl}  & \cellcolor[HTML]{F8FBFF}25.38          & \multicolumn{1}{c|}{{\color[HTML]{939393} 103.80}} & {\color[HTML]{939393} 26.68}           & \multicolumn{1}{c|}{38.44}                         & {\color[HTML]{939393} 27.02}           & \multicolumn{1}{c|}{41.59}                         & {\color[HTML]{939393} 27.57} & \multicolumn{1}{c|}{29.68}         & {\color[HTML]{939393} 24.12} & \multicolumn{1}{c|}{27.76}         & {\color[HTML]{939393} 23.48} & 27.36         \\
\multicolumn{1}{c|}{MACE}    & \cellcolor[HTML]{F8FBFF}{\ul 20.78}    & \multicolumn{1}{c|}{{\color[HTML]{939393} 169.22}} & {\color[HTML]{939393} 22.95}           & \multicolumn{1}{c|}{118.01}                        & {\color[HTML]{939393} 23.33}           & \multicolumn{1}{c|}{111.90}                        & {\color[HTML]{939393} 25.77} & \multicolumn{1}{c|}{81.99}         & {\color[HTML]{939393} 23.96} & \multicolumn{1}{c|}{43.27}         & {\color[HTML]{939393} 22.25} & 65.97         \\
\multicolumn{1}{c|}{SPM}     & \cellcolor[HTML]{F8FBFF}23.89          & \multicolumn{1}{c|}{{\color[HTML]{939393} 122.63}} & {\color[HTML]{939393} 26.66}           & \multicolumn{1}{c|}{{\ul 33.06}}                   & {\color[HTML]{939393} 27.12}           & \multicolumn{1}{c|}{{\ul 34.70}}                   & {\color[HTML]{939393} 27.51} & \multicolumn{1}{c|}{{\ul 23.89}}   & {\color[HTML]{939393} 24.24} & \multicolumn{1}{c|}{{\ul 19.61}}   & {\color[HTML]{939393} 23.70} & {\ul 18.26}   \\
\multicolumn{1}{c|}{NP}      & \cellcolor[HTML]{F8FBFF}23.66          & \multicolumn{1}{c|}{{\color[HTML]{939393} 125.98}} & {\color[HTML]{939393} 26.14}           & \multicolumn{1}{c|}{59.58}                         & {\color[HTML]{939393} 26.66}           & \multicolumn{1}{c|}{78.74}                         & {\color[HTML]{939393} 27.36} & \multicolumn{1}{c|}{52.37}         & {\color[HTML]{939393} 23.89} & \multicolumn{1}{c|}{67.51}         & {\color[HTML]{939393} 22.16} & 55.22         \\
\multicolumn{1}{c|}{SLD}     & \cellcolor[HTML]{F8FBFF}27.84          & \multicolumn{1}{c|}{{\color[HTML]{939393} 64.78}}  & {\color[HTML]{939393} 26.46}           & \multicolumn{1}{c|}{48.12}                         & {\color[HTML]{939393} 27.52}           & \multicolumn{1}{c|}{55.36}                         & {\color[HTML]{939393} 27.33} & \multicolumn{1}{c|}{38.74}         & {\color[HTML]{939393} 24.03} & \multicolumn{1}{c|}{41.95}         & {\color[HTML]{939393} 22.80} & 49.08         \\ 
\multicolumn{1}{c|}{Ours}    & \cellcolor[HTML]{F8FBFF}\textbf{20.28} & \multicolumn{1}{c|}{{\color[HTML]{939393} 120.46}} & {\color[HTML]{939393} 26.53}           & \multicolumn{1}{c|}{\textbf{5.72}}                 & {\color[HTML]{939393} 27.25}           & \multicolumn{1}{c|}{\textbf{8.56}}                 & {\color[HTML]{939393} 27.40} & \multicolumn{1}{c|}{\textbf{5.79}} & {\color[HTML]{939393} 24.27} & \multicolumn{1}{c|}{\textbf{2.32}} & {\color[HTML]{939393} 23.77} & \textbf{6.07} \\ \hline
                             & \multicolumn{12}{c}{\textit{Erase \textbf{Snoopy} and \textbf{Mickey}}}                                                                                                                                                                                                                                                                                                                                                                                                                               \\ \hline
\multicolumn{1}{c|}{}        & \cellcolor[HTML]{F8FBFF}CS $\downarrow$            & \multicolumn{1}{c|}{{\color[HTML]{939393} FID}}    & \cellcolor[HTML]{F8FBFF}CS $\downarrow$             & \multicolumn{1}{c|}{{\color[HTML]{939393} FID}}    & {\color[HTML]{939393} CS} $\downarrow$             & \multicolumn{1}{c|}{FID $\downarrow$}                           & {\color[HTML]{939393} CS}    & \multicolumn{1}{c|}{FID $\downarrow$}           & {\color[HTML]{939393} CS}    & \multicolumn{1}{c|}{FID $\downarrow$}           & {\color[HTML]{939393} CS}    & FID $\downarrow$          \\ \hline
\multicolumn{1}{c|}{ConAbl}  & \cellcolor[HTML]{F8FBFF}24.26          & \multicolumn{1}{c|}{{\color[HTML]{939393} 119.96}} & \cellcolor[HTML]{F8FBFF}24.08          & \multicolumn{1}{c|}{{\color[HTML]{939393} 96.94}}  & {\color[HTML]{939393} 27.02}           & \multicolumn{1}{c|}{46.32}                         & {\color[HTML]{939393} 27.75} & \multicolumn{1}{c|}{39.63}         & {\color[HTML]{939393} 23.98} & \multicolumn{1}{c|}{30.57}         & {\color[HTML]{939393} 23.33} & 27.49         \\
\multicolumn{1}{c|}{MACE}    & \cellcolor[HTML]{F8FBFF}{\ul 20.74}    & \multicolumn{1}{c|}{{\color[HTML]{939393} 171.16}} & \cellcolor[HTML]{F8FBFF}{\ul 20.71}    & \multicolumn{1}{c|}{{\color[HTML]{939393} 140.50}} & {\color[HTML]{939393} 25.87}           & \multicolumn{1}{c|}{51.49}                         & {\color[HTML]{939393} 25.87} & \multicolumn{1}{c|}{110.67}        & {\color[HTML]{939393} 23.82} & \multicolumn{1}{c|}{52.07}         & {\color[HTML]{939393} 21.70} & 77.13         \\
\multicolumn{1}{c|}{SPM}     & \cellcolor[HTML]{F8FBFF}23.16          & \multicolumn{1}{c|}{{\color[HTML]{939393} 128.08}} & \cellcolor[HTML]{F8FBFF}22.81          & \multicolumn{1}{c|}{{\color[HTML]{939393} 115.02}} & {\color[HTML]{939393} 26.92}           & \multicolumn{1}{c|}{{\ul 41.58}}                   & {\color[HTML]{939393} 27.45} & \multicolumn{1}{c|}{{\ul 31.77}}   & {\color[HTML]{939393} 24.13} & \multicolumn{1}{c|}{{\ul 21.96}}   & {\color[HTML]{939393} 23.60} & {\ul 23.69}   \\
\multicolumn{1}{c|}{NP}      & \cellcolor[HTML]{F8FBFF}23.59          & \multicolumn{1}{c|}{{\color[HTML]{939393} 124.10}} & \cellcolor[HTML]{F8FBFF}24.85          & \multicolumn{1}{c|}{{\color[HTML]{939393} 83.68}}  & {\color[HTML]{939393} 26.69}           & \multicolumn{1}{c|}{81.41}                         & {\color[HTML]{939393} 27.27} & \multicolumn{1}{c|}{50.10}         & {\color[HTML]{939393} 23.62} & \multicolumn{1}{c|}{65.93}         & {\color[HTML]{939393} 21.84} & 58.88         \\
\multicolumn{1}{c|}{SLD}     & \cellcolor[HTML]{F8FBFF}27.76          & \multicolumn{1}{c|}{{\color[HTML]{939393} 59.97}}  & \cellcolor[HTML]{F8FBFF}26.74          & \multicolumn{1}{c|}{{\color[HTML]{939393} 50.16}}  & {\color[HTML]{939393} 27.53}           & \multicolumn{1}{c|}{54.59}                         & {\color[HTML]{939393} 27.29} & \multicolumn{1}{c|}{39.24}         & {\color[HTML]{939393} 23.97} & \multicolumn{1}{c|}{41.62}         & {\color[HTML]{939393} 22.66} & 50.13         \\ 
\multicolumn{1}{c|}{Ours}    & \cellcolor[HTML]{F8FBFF}\textbf{20.29} & \multicolumn{1}{c|}{{\color[HTML]{939393} 121.12}} & \cellcolor[HTML]{F8FBFF}\textbf{19.93} & \multicolumn{1}{c|}{{\color[HTML]{939393} 108.22}} & {\color[HTML]{939393} 27.27}           & \multicolumn{1}{c|}{\textbf{9.34}}                 & {\color[HTML]{939393} 27.42} & \multicolumn{1}{c|}{\textbf{5.84}} & {\color[HTML]{939393} 24.26} & \multicolumn{1}{c|}{\textbf{2.41}} & {\color[HTML]{939393} 23.73} & \textbf{6.43} \\ \hline
\multicolumn{1}{l}{}         & \multicolumn{12}{c}{\textit{Erase \textbf{Snoopy} and \textbf{Mickey} and \textbf{Spongebob}}}                                                                                                                                                                                                                                                                                                                                                                                                               \\ \hline
\multicolumn{1}{l|}{}        & \cellcolor[HTML]{F8FBFF}CS $\downarrow$            & \multicolumn{1}{c|}{{\color[HTML]{939393} FID}}    & \cellcolor[HTML]{F8FBFF}CS  $\downarrow$           & \multicolumn{1}{c|}{{\color[HTML]{939393} FID}}    & \cellcolor[HTML]{F8FBFF}CS $\downarrow$             & \multicolumn{1}{c|}{{\color[HTML]{939393} FID}}    & {\color[HTML]{939393} CS}    & \multicolumn{1}{c|}{FID $\downarrow$}           & {\color[HTML]{939393} CS}    & \multicolumn{1}{c|}{FID $\downarrow$}           & {\color[HTML]{939393} CS}    & FID $\downarrow$          \\ \hline
\multicolumn{1}{c|}{ConAbl}  & \cellcolor[HTML]{F8FBFF}23.94          & \multicolumn{1}{c|}{{\color[HTML]{939393} 126.70}} & \cellcolor[HTML]{F8FBFF}23.64          & \multicolumn{1}{c|}{{\color[HTML]{939393} 105.07}} & \cellcolor[HTML]{F8FBFF}25.04          & \multicolumn{1}{c|}{{\color[HTML]{939393} 108.67}} & {\color[HTML]{939393} 27.76} & \multicolumn{1}{c|}{51.20}         & {\color[HTML]{939393} 23.83} & \multicolumn{1}{c|}{23.83}         & {\color[HTML]{939393} 23.17} & 30.03         \\
\multicolumn{1}{c|}{MACE}    & \cellcolor[HTML]{F8FBFF}{\ul 20.48}    & \multicolumn{1}{c|}{{\color[HTML]{939393} 172.80}} & \cellcolor[HTML]{F8FBFF}{\ul 20.50}    & \multicolumn{1}{c|}{{\color[HTML]{939393} 143.66}} & \cellcolor[HTML]{F8FBFF}21.59          & \multicolumn{1}{c|}{{\color[HTML]{939393} 120.87}} & {\color[HTML]{939393} 24.38} & \multicolumn{1}{c|}{99.68}         & {\color[HTML]{939393} 23.70} & \multicolumn{1}{c|}{47.46}         & {\color[HTML]{939393} 21.74} & 70.38         \\
\multicolumn{1}{c|}{SPM}     & \cellcolor[HTML]{F8FBFF}22.81          & \multicolumn{1}{c|}{{\color[HTML]{939393} 133.06}} & \cellcolor[HTML]{F8FBFF}22.35          & \multicolumn{1}{c|}{{\color[HTML]{939393} 121.85}} & \cellcolor[HTML]{F8FBFF}{\ul 20.82}    & \multicolumn{1}{c|}{{\color[HTML]{939393} 152.72}} & {\color[HTML]{939393} 27.45} & \multicolumn{1}{c|}{39.83}         & {\color[HTML]{939393} 24.10} & \multicolumn{1}{c|}{{\ul 22.68}}   & {\color[HTML]{939393} 23.52} & {\ul 25.31}   \\
\multicolumn{1}{c|}{NP}      & \cellcolor[HTML]{F8FBFF}24.29          & \multicolumn{1}{c|}{{\color[HTML]{939393} 129.75}} & \cellcolor[HTML]{F8FBFF}24.76          & \multicolumn{1}{c|}{{\color[HTML]{939393} 89.74}}  & \cellcolor[HTML]{F8FBFF}25.31          & \multicolumn{1}{c|}{{\color[HTML]{939393} 106.30}} & {\color[HTML]{939393} 27.28} & \multicolumn{1}{c|}{64.75}         & {\color[HTML]{939393} 23.55} & \multicolumn{1}{c|}{65.10}         & {\color[HTML]{939393} 21.63} & 59.33         \\
\multicolumn{1}{c|}{SLD}     & \cellcolor[HTML]{F8FBFF}27.84          & \multicolumn{1}{c|}{{\color[HTML]{939393} 58.16}}  & \cellcolor[HTML]{F8FBFF}26.71          & \multicolumn{1}{c|}{{\color[HTML]{939393} 49.70}}  & \cellcolor[HTML]{F8FBFF}27.60          & \multicolumn{1}{c|}{{\color[HTML]{939393} 54.61}}  & {\color[HTML]{939393} 27.35} & \multicolumn{1}{c|}{{\ul 39.41}}   & {\color[HTML]{939393} 23.90} & \multicolumn{1}{c|}{42.32}         & {\color[HTML]{939393} 22.46} & 49.88         \\ 
\multicolumn{1}{c|}{Ours}    & \cellcolor[HTML]{F8FBFF}\textbf{19.39} & \multicolumn{1}{c|}{{\color[HTML]{939393} 124.49}} & \cellcolor[HTML]{F8FBFF}\textbf{19.73} & \multicolumn{1}{c|}{{\color[HTML]{939393} 112.97}} & \cellcolor[HTML]{F8FBFF}\textbf{20.34} & \multicolumn{1}{c|}{{\color[HTML]{939393} 118.47}} & {\color[HTML]{939393} 27.42} & \multicolumn{1}{c|}{\textbf{6.85}} & {\color[HTML]{939393} 24.27} & \multicolumn{1}{c|}{\textbf{2.79}} & {\color[HTML]{939393} 23.76} & \textbf{7.26} \\ \hline
\end{tabular}}
\vspace{-2mm}
\caption{\textbf{Extended quantitative comparison of single- and multi-instance erasure.}
The best and second-best results are marked in \textbf{bold} and \underline{underlined}, respectively. Columns in \textcolor[HTML]{939393}{gray} indicate items that do not directly reflect erasure efficacy or prior preservation performance.}
\vspace{-3mm}
\label{tab:instance_extend}
\end{table*}

\begin{table*}[t]
\centering
\footnotesize  
\renewcommand{\arraystretch}{1.20}  
\resizebox{0.70\hsize}{!}{\begin{tabular}{ccccccccccc}
\hline
\multicolumn{1}{c|}{}        & \multicolumn{2}{c|}{Van Gogh}                                                               & \multicolumn{2}{c|}{Picasso}                                                                & \multicolumn{2}{c|}{Monet}                                                                  & \multicolumn{2}{c}{Andy Warhol}                                   & \multicolumn{2}{c}{Caravaggio}               \\ \hline
\multicolumn{1}{c|}{}        & CS           & \multicolumn{1}{c|}{FID}                           & CS                                     & \multicolumn{1}{c|}{FID}                           & CS                                     & \multicolumn{1}{c|}{FID}                           & CS                           & \multicolumn{1}{c|}{FID}           & CS                           & FID           \\ \hline
\multicolumn{1}{c|}{SD v1.4} &29.20          & \multicolumn{1}{c|}{-}                             & 28.84                                  & \multicolumn{1}{c|}{-}                             & 29.41                                  & \multicolumn{1}{c|}{-}                             & 29.73                        & \multicolumn{1}{c|}{-}             & 27.09                        & -             \\ \hline
\multicolumn{11}{c}{\textit{Erase \textbf{Van Gogh}}}                                                                                                                                                                                                                                                                                                                                                                                                     \\ \hline
\multicolumn{1}{c|}{}        & \cellcolor[HTML]{F8FBFF}CS $\downarrow$            & \multicolumn{1}{c|}{{\color[HTML]{939393} FID}}    & {\color[HTML]{939393} CS}              & \multicolumn{1}{c|}{FID $\downarrow$}                           & {\color[HTML]{939393} CS}              & \multicolumn{1}{c|}{FID $\downarrow$}                           & {\color[HTML]{939393} CS}    & \multicolumn{1}{c|}{FID $\downarrow$}           & {\color[HTML]{939393} CS}    & FID $\downarrow$          \\ \hline
\multicolumn{1}{c|}{ConAbl}  & \cellcolor[HTML]{F8FBFF}28.80          & \multicolumn{1}{c|}{{\color[HTML]{939393} 120.93}} & {\color[HTML]{939393} 28.10}           & \multicolumn{1}{c|}{71.71}                         & {\color[HTML]{939393} 25.99}           & \multicolumn{1}{c|}{138.72}                        & {\color[HTML]{939393} 29.34} & \multicolumn{1}{c|}{70.30}         & {\color[HTML]{939393} 26.83} & 73.10         \\
\multicolumn{1}{c|}{MACE}    & \cellcolor[HTML]{F8FBFF}27.74          & \multicolumn{1}{c|}{{\color[HTML]{939393} 144.75}} & {\color[HTML]{939393} 28.37}           & \multicolumn{1}{c|}{65.77}                         & {\color[HTML]{939393} 29.48}           & \multicolumn{1}{c|}{69.79}                         & {\color[HTML]{939393} 29.30} & \multicolumn{1}{c|}{83.37}         & {\color[HTML]{939393} 27.11} & 75.41         \\
\multicolumn{1}{c|}{SPM}     & \cellcolor[HTML]{F8FBFF}\textbf{24.78} & \multicolumn{1}{c|}{{\color[HTML]{939393} 185.50}} & {\color[HTML]{939393} 28.34}           & \multicolumn{1}{c|}{{\ul 62.25}}                   & {\color[HTML]{939393} 29.34}           & \multicolumn{1}{c|}{{\ul 32.27}}                   & {\color[HTML]{939393} 29.52} & \multicolumn{1}{c|}{{\ul 58.30}}   & {\color[HTML]{939393} 27.01} & {\ul 61.50}   \\
\multicolumn{1}{c|}{NP}      & \cellcolor[HTML]{F8FBFF}24.90          & \multicolumn{1}{c|}{{\color[HTML]{939393} 193.24}} & {\color[HTML]{939393} 25.11}           & \multicolumn{1}{c|}{141.56}                        & {\color[HTML]{939393} 26.08}           & \multicolumn{1}{c|}{124.52}                        & {\color[HTML]{939393} 27.06} & \multicolumn{1}{c|}{127.85}        & {\color[HTML]{939393} 25.34} & 136.32        \\
\multicolumn{1}{c|}{SLD}     & \cellcolor[HTML]{F8FBFF}27.48          & \multicolumn{1}{c|}{{\color[HTML]{939393} 133.07}} & {\color[HTML]{939393} 26.89}           & \multicolumn{1}{c|}{103.96}                        & {\color[HTML]{939393} 27.61}           & \multicolumn{1}{c|}{109.11}                        & {\color[HTML]{939393} 28.24} & \multicolumn{1}{c|}{103.89}        & {\color[HTML]{939393} 25.82} & 119.32        \\
\multicolumn{1}{c|}{SAFREE}  & \cellcolor[HTML]{F8FBFF}25.82          & \multicolumn{1}{c|}{{\color[HTML]{939393} 183.06}} & {\color[HTML]{939393} 25.84}           & \multicolumn{1}{c|}{130.35}                        & {\color[HTML]{939393} 27.15}           & \multicolumn{1}{c|}{128.71}                        & {\color[HTML]{939393} 27.20} & \multicolumn{1}{c|}{127.72}        & {\color[HTML]{939393} 25.53} & 134.46        \\
\multicolumn{1}{c|}{Ours}    & \cellcolor[HTML]{F8FBFF}{\ul 24.87}    & \multicolumn{1}{c|}{{\color[HTML]{939393} 188.94}} & {\color[HTML]{939393} 28.80}           & \multicolumn{1}{c|}{\textbf{6.82}}                 & {\color[HTML]{939393} 29.43}           & \multicolumn{1}{c|}{\textbf{2.66}}                 & {\color[HTML]{939393} 29.74} & \multicolumn{1}{c|}{\textbf{8.36}} & {\color[HTML]{939393} 27.09} & \textbf{6.84} \\ \hline
\multicolumn{11}{c}{\textit{Erase \textbf{Picasso}}}                                                                                                                                                                                                                                                                                                                                                                                                      \\ \hline
\multicolumn{1}{c|}{}        & {\color[HTML]{939393} CS}              & \multicolumn{1}{c|}{FID $\downarrow$}                           & \cellcolor[HTML]{F8FBFF}CS   $\downarrow$          & \multicolumn{1}{c|}{{\color[HTML]{939393} FID}}    & {\color[HTML]{939393} CS}              & \multicolumn{1}{c|}{FID $\downarrow$}                           & {\color[HTML]{939393} CS}    & \multicolumn{1}{c|}{FID $\downarrow$}           & {\color[HTML]{939393} CS}    & FID  $\downarrow$         \\ \hline
\multicolumn{1}{c|}{ConAbl}  & {\color[HTML]{939393} 29.46}           & \multicolumn{1}{c|}{58.62}                         & \cellcolor[HTML]{F8FBFF}27.72          & \multicolumn{1}{c|}{{\color[HTML]{939393} 121.45}} & {\color[HTML]{939393} 26.37}           & \multicolumn{1}{c|}{140.34}                        & {\color[HTML]{939393} 29.51} & \multicolumn{1}{c|}{73.35}         & {\color[HTML]{939393} 27.17} & 67.44         \\
\multicolumn{1}{c|}{MACE}    & {\color[HTML]{939393} 29.73}           & \multicolumn{1}{c|}{60.46}                         & \cellcolor[HTML]{F8FBFF}27.11          & \multicolumn{1}{c|}{{\color[HTML]{939393} 131.82}} & {\color[HTML]{939393} 29.44}           & \multicolumn{1}{c|}{49.92}                         & {\color[HTML]{939393} 29.65} & \multicolumn{1}{c|}{76.10}         & {\color[HTML]{939393} 27.08} & 72.85         \\
\multicolumn{1}{c|}{SPM}     & {\color[HTML]{939393} 29.26}           & \multicolumn{1}{c|}{{\ul 38.79}}                   & \cellcolor[HTML]{F8FBFF}{\ul 26.69}    & \multicolumn{1}{c|}{{\color[HTML]{939393} 157.32}} & {\color[HTML]{939393} 29.44}           & \multicolumn{1}{c|}{{\ul 7.76}}                    & {\color[HTML]{939393} 29.67} & \multicolumn{1}{c|}{{\ul 52.00}}   & {\color[HTML]{939393} 27.08} & {\ul 51.40}   \\
\multicolumn{1}{c|}{NP}      & {\color[HTML]{939393} 29.28}           & \multicolumn{1}{c|}{111.35}                        & \cellcolor[HTML]{F8FBFF}\textbf{26.14} & \multicolumn{1}{c|}{{\color[HTML]{939393} 169.23}} & {\color[HTML]{939393} 29.34}           & \multicolumn{1}{c|}{91.11}                         & {\color[HTML]{939393} 28.14} & \multicolumn{1}{c|}{116.24}        & {\color[HTML]{939393} 26.50} & 121.82        \\
\multicolumn{1}{c|}{SLD}     & {\color[HTML]{939393} 29.36}           & \multicolumn{1}{c|}{98.21}                         & \cellcolor[HTML]{F8FBFF}27.03          & \multicolumn{1}{c|}{{\color[HTML]{939393} 105.37}} & {\color[HTML]{939393} 29.79}           & \multicolumn{1}{c|}{93.01}                         & {\color[HTML]{939393} 28.80} & \multicolumn{1}{c|}{97.00}         & {\color[HTML]{939393} 26.42} & 110.05        \\
\multicolumn{1}{c|}{SAFREE}  & {\color[HTML]{939393} 29.96}           & \multicolumn{1}{c|}{117.32}                        & \cellcolor[HTML]{F8FBFF}26.42          & \multicolumn{1}{c|}{{\color[HTML]{939393} 183.80}} & {\color[HTML]{939393} 29.45}           & \multicolumn{1}{c|}{93.51}                        & {\color[HTML]{939393} 27.88} & \multicolumn{1}{c|}{122.89}        & {\color[HTML]{939393} 26.32} & 116.51        \\
\multicolumn{1}{c|}{Ours}    & {\color[HTML]{939393} 29.17}           & \multicolumn{1}{c|}{\textbf{5.49}}                                      & \cellcolor[HTML]{F8FBFF}26.99          & \multicolumn{1}{c|}{{\color[HTML]{939393} 132.64}} & {\color[HTML]{939393} 29.43}           & \multicolumn{1}{c|}{\textbf{2.33}}                 & {\color[HTML]{939393} 29.72} & \multicolumn{1}{c|}{\textbf{9.38}} & {\color[HTML]{939393} 27.09} & \textbf{7.05} \\ \hline
\multicolumn{11}{c}{\textit{Erase \textbf{Monet}}}                                                                                                                                                                                                                                                                                                                                                                                                        \\ \hline
\multicolumn{1}{c|}{}        & {\color[HTML]{939393} CS}              & \multicolumn{1}{c|}{FID $\downarrow$}                           & {\color[HTML]{939393} CS}              & \multicolumn{1}{c|}{FID $\downarrow$}                           & \cellcolor[HTML]{F8FBFF}CS   $\downarrow$          & \multicolumn{1}{c|}{{\color[HTML]{939393} FID}}    & {\color[HTML]{939393} CS}    & \multicolumn{1}{c|}{FID $\downarrow$}           & {\color[HTML]{939393} CS}    & FID  $\downarrow$        \\ \hline
\multicolumn{1}{c|}{ConAbl}  & {\color[HTML]{939393} 25.84}           & \multicolumn{1}{c|}{141.52}                        & {\color[HTML]{939393} 25.47}           & \multicolumn{1}{c|}{132.10}                        & \cellcolor[HTML]{F8FBFF}{\ul 24.53}    & \multicolumn{1}{c|}{{\color[HTML]{939393} 143.48}} & {\color[HTML]{939393} 26.25} & \multicolumn{1}{c|}{208.38}        & {\color[HTML]{939393} 25.48} & 186.26        \\
\multicolumn{1}{c|}{MACE}    & {\color[HTML]{939393} 29.47}           & \multicolumn{1}{c|}{76.90}                         & {\color[HTML]{939393} 28.56}           & \multicolumn{1}{c|}{69.35}                         & \cellcolor[HTML]{F8FBFF}26.89          & \multicolumn{1}{c|}{{\color[HTML]{939393} 109.58}} & {\color[HTML]{939393} 29.34} & \multicolumn{1}{c|}{88.35}         & {\color[HTML]{939393} 26.75} & 81.72         \\
\multicolumn{1}{c|}{SPM}     & {\color[HTML]{939393} 29.19}           & \multicolumn{1}{c|}{{\ul 41.03}}                   & {\color[HTML]{939393} 28.65}           & \multicolumn{1}{c|}{{\ul 29.71}}                   & \cellcolor[HTML]{F8FBFF}27.00          & \multicolumn{1}{c|}{{\color[HTML]{939393} 105.09}} & {\color[HTML]{939393} 29.65} & \multicolumn{1}{c|}{{\ul 31.90}}   & {\color[HTML]{939393} 29.65} & {\ul 25.99}   \\
\multicolumn{1}{c|}{NP}      & {\color[HTML]{939393} 26.31}           & \multicolumn{1}{c|}{137.21}                        & {\color[HTML]{939393} 25.59}           & \multicolumn{1}{c|}{126.75}                        & \cellcolor[HTML]{F8FBFF}\textbf{24.47} & \multicolumn{1}{c|}{{\color[HTML]{939393} 140.92}} & {\color[HTML]{939393} 27.05} & \multicolumn{1}{c|}{127.22}        & {\color[HTML]{939393} 24.85} & 135.83        \\
\multicolumn{1}{c|}{SLD}     & {\color[HTML]{939393} 28.22}           & \multicolumn{1}{c|}{94.48}                         & {\color[HTML]{939393} 27.10}           & \multicolumn{1}{c|}{92.88}                         & \cellcolor[HTML]{F8FBFF}25.73          & \multicolumn{1}{c|}{{\color[HTML]{939393} 120.14}} & {\color[HTML]{939393} 28.34} & \multicolumn{1}{c|}{100.90}        & {\color[HTML]{939393} 25.45} & 114.87        \\
\multicolumn{1}{c|}{SAFREE}  & {\color[HTML]{939393} 26.07}           & \multicolumn{1}{c|}{125.98}                        & {\color[HTML]{939393} 26.25}           & \multicolumn{1}{c|}{119.19}                        & \cellcolor[HTML]{F8FBFF}25.33          & \multicolumn{1}{c|}{{\color[HTML]{939393} 153.96}} & {\color[HTML]{939393} 26.82} & \multicolumn{1}{c|}{125.27}        & {\color[HTML]{939393} 25.45} & 129.07       \\
\multicolumn{1}{c|}{Ours}    & {\color[HTML]{939393} 29.19}           & \multicolumn{1}{c|}{\textbf{6.94}}                 & {\color[HTML]{939393} 28.80}           & \multicolumn{1}{c|}{\textbf{6.50}}                 & \cellcolor[HTML]{F8FBFF}26.30          & \multicolumn{1}{c|}{{\color[HTML]{939393} 114.06}} & {\color[HTML]{939393} 29.76} & \multicolumn{1}{c|}{\textbf{8.46}} & {\color[HTML]{939393} 27.10} & \textbf{7.19} \\ \hline
\end{tabular}}
\vspace{-2mm}
\caption{\textbf{Extended quantitative comparison of art style erasure.} AdaVD achieves a superior balance between erasure efficacy and prior preservation, especially excelling in prior preservation. 
Notably, it outperforms the concurrent method SAFREE, which also employs orthogonal decomposition.}
\label{tab:style_extend}
\end{table*}

\subsection{On Celebrity Erasure} \label{sec:celebrity}

We experiment with erasing different celebrity concepts, including
\textit{``Bruce Lee''}, \textit{``Marilyn Monroe''}, and \textit{``Melania Trump''}. Five types of prompts were tested,  each containing a distinct concept from \textit{``Bruce Lee''}, \textit{``Marilyn Monroe''}, \textit{``Melania Trump''}, \textit{``Anne Hathaway''} and \textit{``Tom Cruise''}.
As reported in Table \ref{tab:cele}, AdaVD consistently exhibits superior erasing efficacy with prior preservation. 
When erasing different celebrities, AdaVD achieves the lowest or near-lowest CS and FID values, particularly excelling in FID. 
Although SPM ranks the second in prior preservation based on its FID scores, it falls significantly behind in its overall prior preservation quality, as compared to AdaVD.

\begin{figure*}[!t]
\centering
\includegraphics[width=\textwidth]{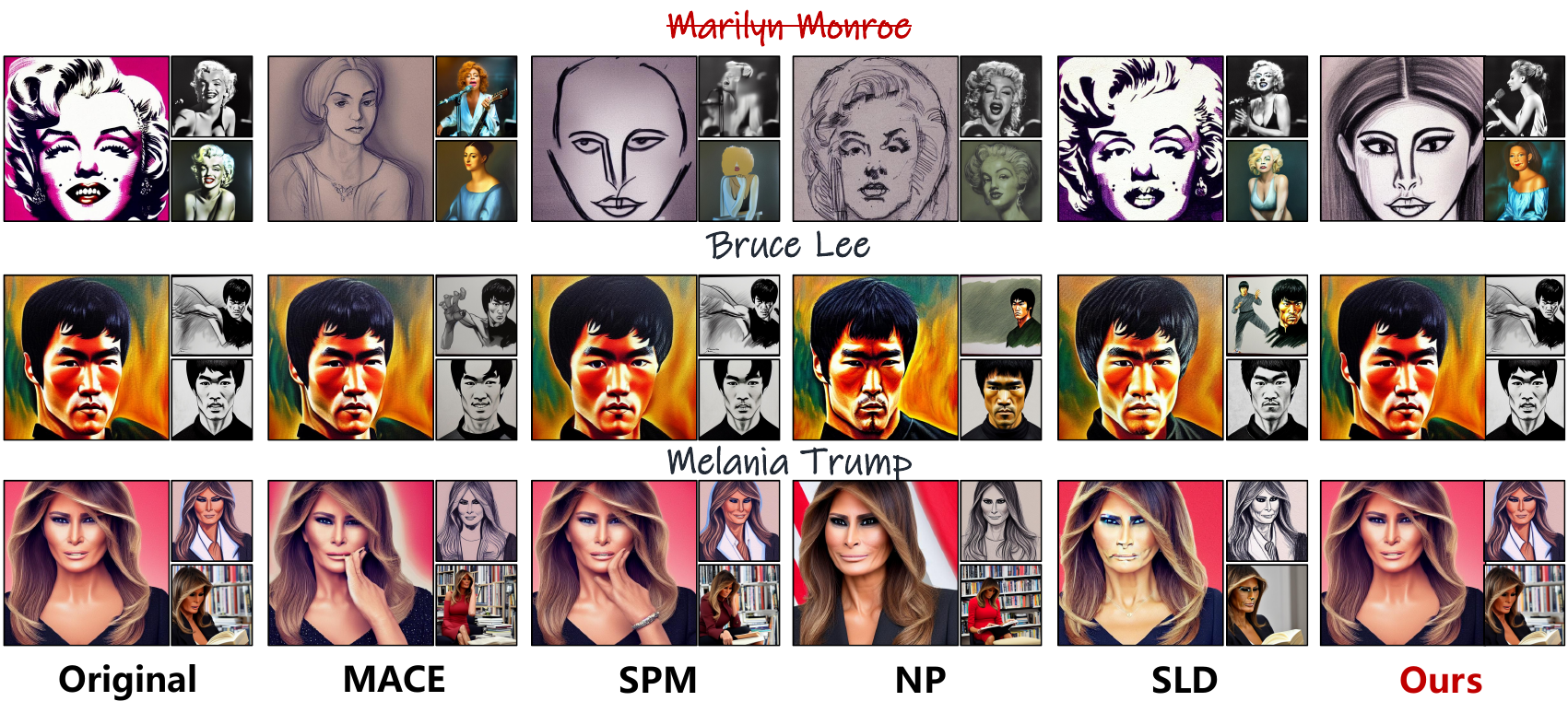}
\caption{\textbf{Qualitative comparison of celebrity erasure.} Our AdaVD can effectively remove the target concept \textit{``Marilyn Monroe"} while preserving non-target celebrities like \textit{``Bruce Lee"} and \textit{``Melania Trump"}.}
\label{fig:celeb}
\end{figure*}

Fig.~\ref{fig:celeb} illustrates and compares generated images of methods, where consistent superior performance of AdaVD can be observed. 
For the target concept \textit{``Marilyn Monroe''},  AdaVD, SPM, and MACE can all successfully remove the celebrity identity.
But SPM is overly aggressive at erasing, obscuring the facial outlines. 
For non-target concepts, all the four competing methods have caused some quite strong deviations, altering the original images.
This is particularly noticeable in the generated images from the prompt corresponding to  \textit{``Melania Trump''}. 
For instance, MACE and SPM have introduced an additional arm in the left image, NP has altered the original pose, and SLD has caused a severe visual change in the mouth and eye areas. 
In contrast, AdaVD is able to successfully maintain all the non-target images with minimal visual changes.
\begin{table}[t]
\centering
\footnotesize  
\renewcommand{\arraystretch}{1.15}  
\resizebox{0.70\hsize}{!}{\begin{tabular}{ccccccccccc}
\hline
\multicolumn{1}{c|}{}        & \multicolumn{2}{c|}{Bruce Lee}                                                              & \multicolumn{2}{c|}{Marilyn Monroe}                                                         & \multicolumn{2}{c|}{Melania Trump}                                                          & \multicolumn{2}{c|}{Anne Hathaway}                                & \multicolumn{2}{c}{Tom Cruise}                \\ \hline
\multicolumn{1}{c|}{}        & CS                                     & \multicolumn{1}{c|}{FID}                           & CS                                     & \multicolumn{1}{c|}{FID}                           & CS                                     & \multicolumn{1}{c|}{FID}                           & CS                           & \multicolumn{1}{c|}{FID}           & CS                           & FID            \\ \hline
\multicolumn{1}{c|}{SD v1.4} & 30.77                                  & \multicolumn{1}{c|}{-}                             & 27.70                                  & \multicolumn{1}{c|}{-}                             & 29.80                                  & \multicolumn{1}{c|}{-}                             & 31.96                        & \multicolumn{1}{c|}{-}             & 31.12                        & -              \\ \hline
\multicolumn{11}{c}{\textit{Erase \textbf{Bruce Lee}}}                                                                                                                                                                                                                                                                                                                                                                                                             \\ \hline
\multicolumn{1}{c|}{}        & \cellcolor[HTML]{F8FBFF}CS $\downarrow$            & \multicolumn{1}{c|}{{\color[HTML]{939393} FID}}    & {\color[HTML]{939393} CS}              & \multicolumn{1}{c|}{FID $\downarrow$}                           & {\color[HTML]{939393} CS}              & \multicolumn{1}{c|}{FID $\downarrow$}                           & {\color[HTML]{939393} CS}    & \multicolumn{1}{c|}{FID $\downarrow$}           & {\color[HTML]{939393} CS}    & FID $\downarrow$          \\ \hline
\multicolumn{1}{c|}{ConAbl}  & \cellcolor[HTML]{F8FBFF}31.35          & \multicolumn{1}{c|}{{\color[HTML]{939393} 87.57}}  & {\color[HTML]{939393} 28.23}           & \multicolumn{1}{c|}{57.79}                         & {\color[HTML]{939393} 29.77}           & \multicolumn{1}{c|}{40.95}                         & {\color[HTML]{939393} 29.77} & \multicolumn{1}{c|}{40.95}         & {\color[HTML]{939393} 30.97} & 53.53          \\
\multicolumn{1}{c|}{MACE}    & \cellcolor[HTML]{F8FBFF}25.04          & \multicolumn{1}{c|}{{\color[HTML]{939393} 131.29}} & {\color[HTML]{939393} 28.13}           & \multicolumn{1}{c|}{74.80}                         & {\color[HTML]{939393} 30.07}           & \multicolumn{1}{c|}{68.83}                         & {\color[HTML]{939393} 31.91} & \multicolumn{1}{c|}{75.05}         & {\color[HTML]{939393} 28.13} & 71.20          \\
\multicolumn{1}{c|}{SPM}     & \cellcolor[HTML]{F8FBFF}27.75          & \multicolumn{1}{c|}{{\color[HTML]{939393} 123.67}} & {\color[HTML]{939393} 27.71}           & \multicolumn{1}{c|}{{\ul 26.89}}                   & {\color[HTML]{939393} 29.81}           & \multicolumn{1}{c|}{{\ul 7.83}}                    & {\color[HTML]{939393} 31.96} & \multicolumn{1}{c|}{{\ul 9.46}}    & {\color[HTML]{939393} 31.13} & {\ul 28.54}    \\
\multicolumn{1}{c|}{NP}      & \cellcolor[HTML]{F8FBFF}{\ul 24.70}    & \multicolumn{1}{c|}{{\color[HTML]{939393} 150.85}} & {\color[HTML]{939393} 26.84}           & \multicolumn{1}{c|}{102.67}                        & {\color[HTML]{939393} 28.94}           & \multicolumn{1}{c|}{82.13}                         & {\color[HTML]{939393} 30.34} & \multicolumn{1}{c|}{89.60}         & {\color[HTML]{939393} 29.67} & 89.92          \\
\multicolumn{1}{c|}{SLD}     & \cellcolor[HTML]{F8FBFF}28.22          & \multicolumn{1}{c|}{{\color[HTML]{939393} 102.26}} & {\color[HTML]{939393} 26.29}           & \multicolumn{1}{c|}{87.15}                         & {\color[HTML]{939393} 29.43}           & \multicolumn{1}{c|}{84.32}                         & {\color[HTML]{939393} 30.97} & \multicolumn{1}{c|}{85.37}         & {\color[HTML]{939393} 29.32} & 94.07          \\ 
\multicolumn{1}{c|}{Ours}    & \cellcolor[HTML]{F8FBFF}\textbf{20.67} & \multicolumn{1}{c|}{{\color[HTML]{939393} 138.70}} & {\color[HTML]{939393} 27.70}           & \multicolumn{1}{c|}{\textbf{6.68}}                 & 29.82                                  & \multicolumn{1}{c|}{\textbf{5.08}}                 & {\color[HTML]{939393} 31.97} & \multicolumn{1}{c|}{\textbf{6.39}} & {\color[HTML]{939393} 31.10} & \textbf{13.11} \\ \hline
\multicolumn{11}{c}{\textit{Erase \textbf{Marilyn Monroe}}}                                                                                                                                                                                                                                                                                                                                                                                                        \\ \hline
\multicolumn{1}{c|}{}        & {\color[HTML]{939393} CS}              & \multicolumn{1}{c|}{FID $\downarrow$}                           & \cellcolor[HTML]{F8FBFF}CS $\downarrow$            & \multicolumn{1}{c|}{{\color[HTML]{939393} FID}}    & {\color[HTML]{939393} CS}                                     & \multicolumn{1}{c|}{FID $\downarrow$}                           & {\color[HTML]{939393} CS}    & \multicolumn{1}{c|}{FID $\downarrow$}           & {\color[HTML]{939393} CS}    & FID $\downarrow$            \\ \hline
\multicolumn{1}{c|}{ConAbl}  & {\color[HTML]{939393} 30.88}           & \multicolumn{1}{c|}{66.97}                         & \cellcolor[HTML]{F8FBFF}28.75          & \multicolumn{1}{c|}{{\color[HTML]{939393} 88.45}}  & {\color[HTML]{939393} 29.69}           & \multicolumn{1}{c|}{51.52}                         & {\color[HTML]{939393} 32.05} & \multicolumn{1}{c|}{58.57}         & {\color[HTML]{939393} 31.10} & 54.13          \\
\multicolumn{1}{c|}{MACE}    & {\color[HTML]{939393} 31.30}           & \multicolumn{1}{c|}{76.23}                         & \cellcolor[HTML]{F8FBFF}\textbf{19.52} & \multicolumn{1}{c|}{{\color[HTML]{939393} 148.34}} & {\color[HTML]{939393} 31.93}           & \multicolumn{1}{c|}{71.05}                         & {\color[HTML]{939393} 30.16} & \multicolumn{1}{c|}{74.90}         & {\color[HTML]{939393} 31.52} & 73.06          \\
\multicolumn{1}{c|}{SPM}     & {\color[HTML]{939393} 30.76}           & \multicolumn{1}{c|}{{\ul 32.70}}                   & \cellcolor[HTML]{F8FBFF}21.87          & \multicolumn{1}{c|}{{\color[HTML]{939393} 145.81}} & {\color[HTML]{939393} 29.83}           & \multicolumn{1}{c|}{{\ul 25.27}}                   & {\color[HTML]{939393} 31.96} & \multicolumn{1}{c|}{{\ul 22.86}}   & {\color[HTML]{939393} 31.10} & {\ul 19.34}    \\
\multicolumn{1}{c|}{NP}      & {\color[HTML]{939393} 29.50}           & \multicolumn{1}{c|}{113.12}                        & \cellcolor[HTML]{F8FBFF}25.86          & \multicolumn{1}{c|}{{\color[HTML]{939393} 149.95}} & {\color[HTML]{939393} 29.29}           & \multicolumn{1}{c|}{87.27}                         & {\color[HTML]{939393} 29.42} & \multicolumn{1}{c|}{98.86}         & {\color[HTML]{939393} 30.02} & 86.70          \\
\multicolumn{1}{c|}{SLD}     & {\color[HTML]{939393} 29.59}           & \multicolumn{1}{c|}{87.83}                         & \cellcolor[HTML]{F8FBFF}26.70          & \multicolumn{1}{c|}{{\color[HTML]{939393} 98.51}}  & {\color[HTML]{939393} 28.81}           & \multicolumn{1}{c|}{107.42}                        & {\color[HTML]{939393} 29.25} & \multicolumn{1}{c|}{102.13}        & {\color[HTML]{939393} 30.35} & 81.12          \\ 
\multicolumn{1}{c|}{Ours}    & {\color[HTML]{939393} 30.73}           & \multicolumn{1}{c|}{\textbf{7.88}}                 & \cellcolor[HTML]{F8FBFF}{\ul 19.87}    & \multicolumn{1}{c|}{{\color[HTML]{939393} 116.94}} & {\color[HTML]{939393} 29.80}           & \multicolumn{1}{c|}{\textbf{4.46}}                 & {\color[HTML]{939393} 31.93} & \multicolumn{1}{c|}{\textbf{5.43}} & {\color[HTML]{939393} 31.13} & \textbf{9.33}  \\ \hline
\multicolumn{11}{c}{\textit{Erase \textbf{Melania Trump}}}                                                                                                                                                                                                                                                                                                                                                                                                         \\ \hline
\multicolumn{1}{c|}{}        & {\color[HTML]{939393} CS}              & \multicolumn{1}{c|}{FID $\downarrow$}                           & {\color[HTML]{939393} CS}              & \multicolumn{1}{c|}{FID $\downarrow$}                           & \cellcolor[HTML]{F8FBFF}CS $\downarrow$             & \multicolumn{1}{c|}{{\color[HTML]{939393} FID}}    & {\color[HTML]{939393} CS}    & \multicolumn{1}{c|}{FID $\downarrow$}           & {\color[HTML]{939393} CS}    & FID $\downarrow$           \\ \hline
\multicolumn{1}{c|}{ConAbl}  & {\color[HTML]{939393} 30.62}           & \multicolumn{1}{c|}{54.46}                         & {\color[HTML]{939393} 28.14}           & \multicolumn{1}{c|}{59.10}                         & \cellcolor[HTML]{F8FBFF}29.89          & \multicolumn{1}{c|}{{\color[HTML]{939393} 79.04}}  & {\color[HTML]{939393} 31.94} & \multicolumn{1}{c|}{58.65}         & {\color[HTML]{939393} 31.00} & 54.50          \\
\multicolumn{1}{c|}{MACE}    & {\color[HTML]{939393} 31.30}           & \multicolumn{1}{c|}{78.07}                         & {\color[HTML]{939393} 27.84}           & \multicolumn{1}{c|}{71.34}                         & \cellcolor[HTML]{F8FBFF}20.71          & \multicolumn{1}{c|}{{\color[HTML]{939393} 122.42}} & {\color[HTML]{939393} 31.94} & \multicolumn{1}{c|}{73.49}         & {\color[HTML]{939393} 31.41} & 71.09          \\
\multicolumn{1}{c|}{SPM}     & {\color[HTML]{939393} 30.79}           & \multicolumn{1}{c|}{{\ul 14.08}}                   & {\color[HTML]{939393} 27.63}           & \multicolumn{1}{c|}{{\ul 30.40}}                   & \cellcolor[HTML]{F8FBFF}\textbf{23.12} & \multicolumn{1}{c|}{{\color[HTML]{939393} 129.68}} & {\color[HTML]{939393} 31.86} & \multicolumn{1}{c|}{{\ul 28.85}}   & {\color[HTML]{939393} 31.10} & {\ul 22.35}    \\
\multicolumn{1}{c|}{NP}      & {\color[HTML]{939393} 29.38}           & \multicolumn{1}{c|}{115.35}                        & {\color[HTML]{939393} 27.63}           & \multicolumn{1}{c|}{103.83}                        & \cellcolor[HTML]{F8FBFF}23.73          & \multicolumn{1}{c|}{{\color[HTML]{939393} 131.73}} & {\color[HTML]{939393} 28.72} & \multicolumn{1}{c|}{106.04}        & {\color[HTML]{939393} 30.27} & 106.00         \\
\multicolumn{1}{c|}{SLD}     & {\color[HTML]{939393} 29.55}           & \multicolumn{1}{c|}{90.69}                         & {\color[HTML]{939393} 26.24}           & \multicolumn{1}{c|}{93.93}                         & \cellcolor[HTML]{F8FBFF}25.45          & \multicolumn{1}{c|}{{\color[HTML]{939393} 103.52}} & {\color[HTML]{939393} 28.43} & \multicolumn{1}{c|}{104.48}        & {\color[HTML]{939393} 30.47} & 88.31          \\
\multicolumn{1}{c|}{Ours}    & {\color[HTML]{939393} 30.75}           & \multicolumn{1}{c|}{\textbf{7.32}}                 & {\color[HTML]{939393} 27.69}           & \multicolumn{1}{c|}{\textbf{6.86}}                 & \cellcolor[HTML]{F8FBFF}{\ul 23.28}    & \multicolumn{1}{c|}{{\color[HTML]{939393} 96.66}}  & {\color[HTML]{939393} 31.95} & \multicolumn{1}{c|}{\textbf{6.52}} & {\color[HTML]{939393} 31.08} & \textbf{5.74}  \\ \hline
\end{tabular}}
\vspace{-2mm}
\caption{\textbf{Quantitative comparison of celebrity erasure.} Compared to both training-based and training-free methods, AdaVD achieves an optimal balance between erasure efficacy and prior preservation, demonstrating exceptional performance, particularly in prior preservation.}
\vspace{-3mm}
\label{tab:cele}
\end{table}

\subsection{On NSFW Erasure} \label{sec:nsfw}
\begin{figure*}[!t]
\centering
\includegraphics[width=\textwidth]{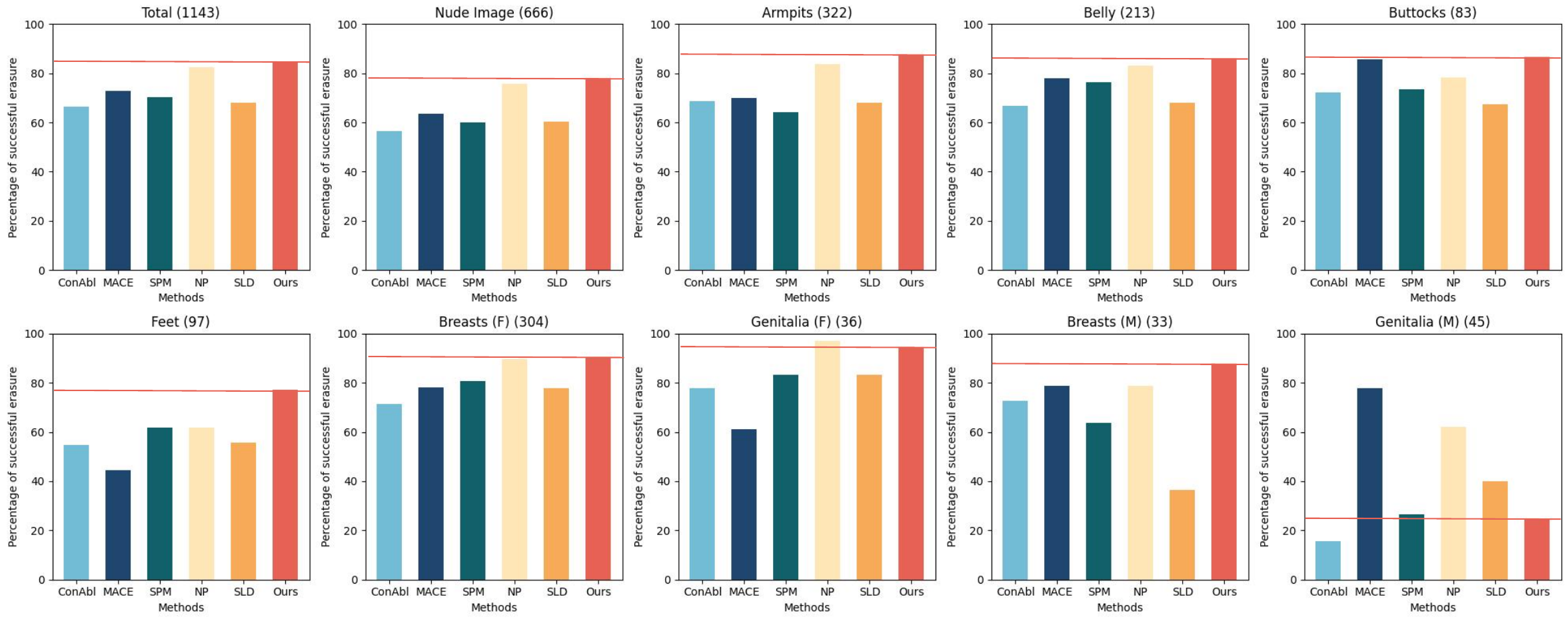}
\caption{\textbf{Performance of AdaVD on NSFW erasure.} The number following each category represents the number of nude items generated by SD v1.4, while each bar illustrates the success rate of erasing the corresponding nude items for each method. Our AdaVD demonstrates superior performance on NSFW erasure, outperforming both training-based and training-free methods.}
\label{fig:nude}
\end{figure*}

Unlike the erasure of specific instances, art styles, and celebrities,   NSFW concept erasure is more challenging.
One reason is that the NSFW concepts are often implicit and hidden within prompts that can be particularly rich in their semantics. 
Also, many NSFW concepts have synonyms, and it is important to remove both the target concept and its synonyms.
For instance, when targeting at removing the \textit{``nudity ''} concept, it is essential to also remove the \textit{``sexual''} concept. 
We experiment with erasing the    \textit{``nudity''} concept using the I2P benchmark.
To examine how well the \textit{``nudity''} concept is erased, we employ the NudeNet with a threshold of 0.3 to detect nudity in the generated images and analyze the total number of nude items and the overall nude images that are detected.

Results are reported in Fig. \ref{fig:nude}, where, 
despite the challenges, AdaVD demonstrates a superior nudity erasure performance, with a semi-threshold and a slower increasing rate.
It outperforms both training-based and training-free methods, achieving the best or close-to-best success rate in nearly all categories, with approximately  85$\%$  of the nude items successfully removed. 
 It is worth mentioning that NudeNet can be overly aggressive at detecting nude items, resulting in detection errors.
 For example, it may incorrectly classify a circle with a dot as \textit{''Female Breast Exposure''} or a person opening their mouth as \textit{''Male Genitalia Exposure''}. 
 We increased the NudeNet threshold to 0.3, in order to mitigate this issue, but still, there is a detection error. 
Being examined by an overly strict nudity detector that can flag sometimes healthy or irrelevant content as nude ones, AdaVD achieves the highest erasure rate for nearly all tested nude items compared to other competing methods, as shown in Fig. \ref{fig:nude}.

\begin{figure*}[!t]
\centering
\includegraphics[width=\textwidth]{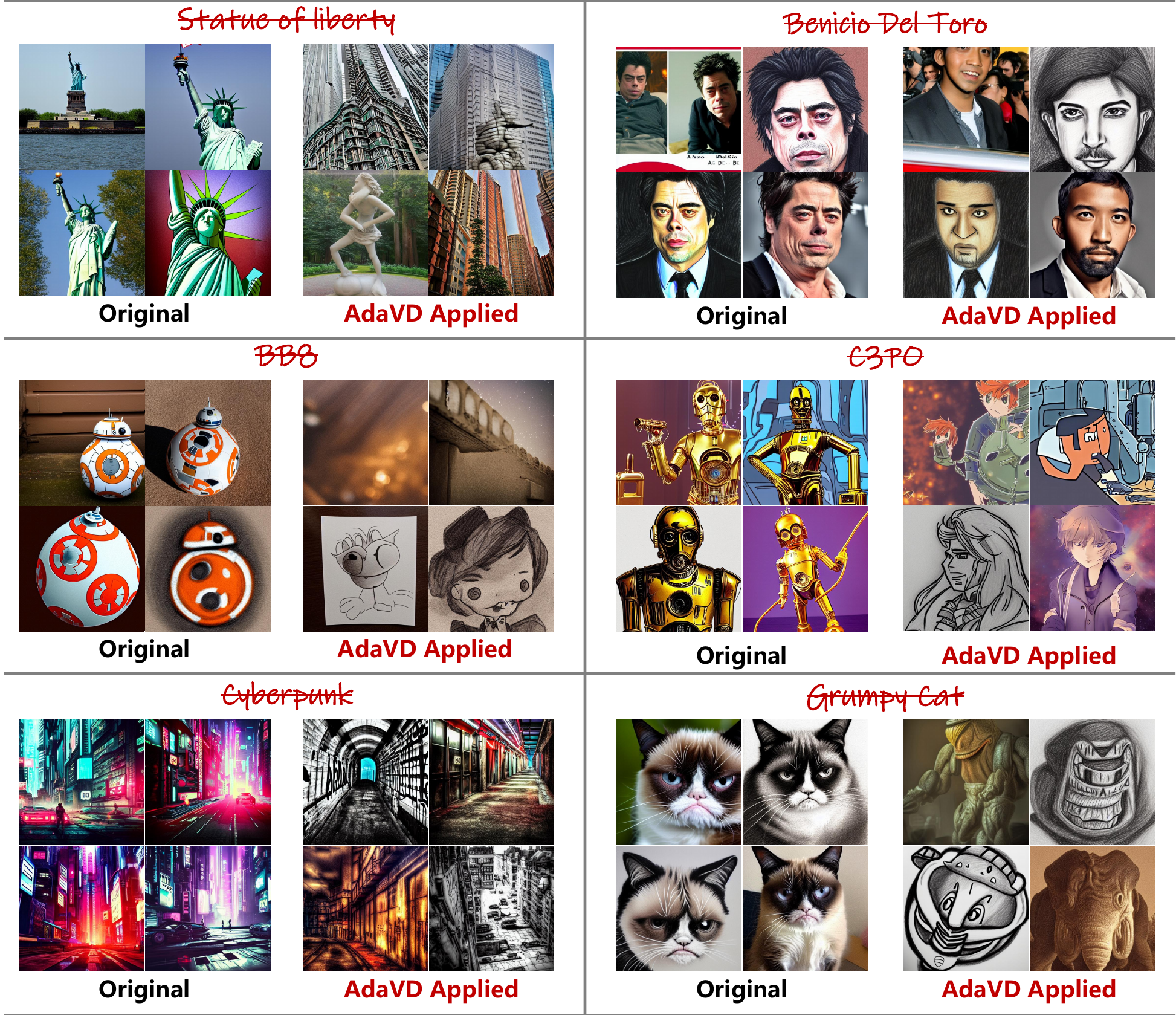}
\caption{\textbf{Extended results of AdaVD in single-concept erasure task.} We present additional generated images after applying AdaVD with SD v1.4 to erase a single concept, further validating the erasure efficacy of our AdaVD.}
\label{fig:appendix-single}
\end{figure*}
\subsection{More Erasure Examples}

We demonstrate additional examples for erasing single concepts from prompts that contain such concepts.
The experimented  concepts include the specific instances of \textit{``Statue of Liberty''}, \textit{``BB8''}, \textit{``C3PO''}, and \textit{``Grumpy Cat''}, the celebrity \textit{``Benicio Del Tor}o'', and the art style \textit{``Cyberpunk''}. 
Among these, \textit{``BB8''} and \textit{``C3PO''} are fictional characters, while \textit{``Statue of Liberty''} and \textit{``Grumpy Cat''} represent realistic entities from daily life.
Fig. \ref{fig:appendix-single} presents the generated image examples.
It can be seen that our AdaVD consistently exhibits superior erasure efficacy across all these concepts, being robust in erasing diverse types of concepts.

\section{On Transferability to Other T2I Models} 
\label{sec:transfer}
The proposed AdaVD is a flexible concept erasure approach that can be transferred to other T2I diffusion models. 
In addition to SD v1.4, as experimented in the main paper, we conduct additional experiments to demonstrate its transferability and effectiveness by integrating it with a series of other T2I diffusion models.

\subsection{AdaVD on SDXL v1.0}
\begin{figure*}[!t]
\centering
\includegraphics[width=\textwidth]{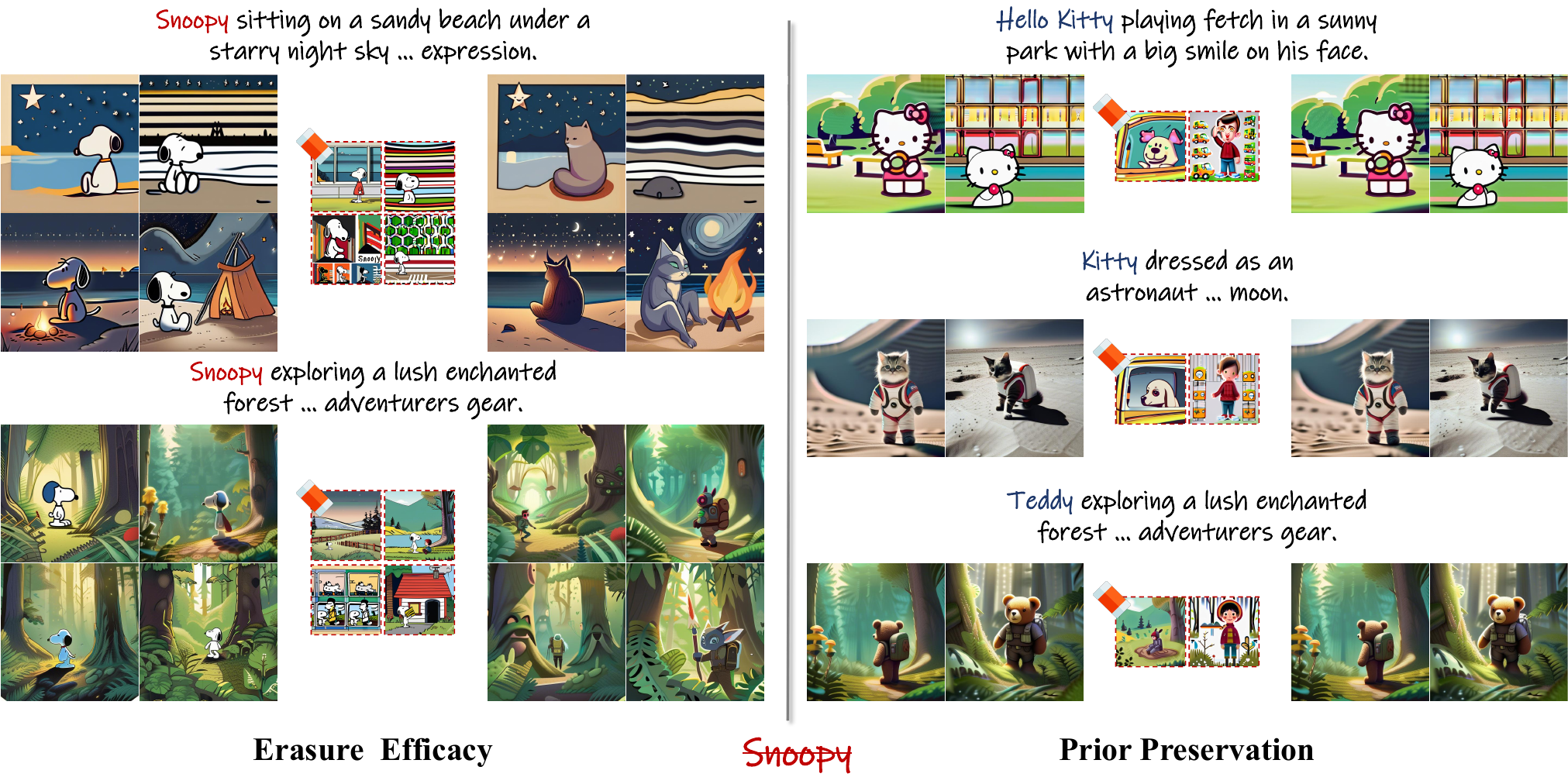}
\caption{\textbf{Results of AdaVD on SDXL v1.0 for erasing \textit{``Snoopy"}}: Our AdaVD effectively supports SDXL v1.0, which has a different structural design than SD v1.4, in achieving effective erasure of the target concept. Additionally, AdaVD demonstrates excellent prior preservation, as evidenced by its ability to generate non-target concepts like \textit{``Hello Kitty"}, \textit{``Kitty"}, and \textit{``Teddy"} even with semantically rich prompts. AdaVD successfully retains nearly all details in non-target content, underscoring its capability for precise erasure without compromising unrelated elements.}
\label{fig:sdxl}
\end{figure*}

We integrate AdaVD with SDXL v1.0 \cite{podellsdxl} which has a different architecture from  SD v1.4-v2.1. 
It employs two distinct text encoders to process textual prompts, and their outputs are concatenated and fed into the CA layers to interact with the latent representations of the noisy images. 
Also, the generated text embeddings are enhanced by time embeddings to ensure the alignment between textual prompts and timesteps. 
Following the same approach as how it is coupled with SD v1.4, AdaVD is applied in the value space at each CA layer within the UNet of SDXL v1.0. 
For the target concepts,  both sets of their embeddings computed by the two text encoders are pre-processed following the procedure outlined in Sec. \ref{sec:pre-processing},  then they are used to start the erasure process following the method outlined in Sec. \ref{sec_singlecon}.

Fig. \ref{fig:sdxl} demonstrates the generated image examples by coupling  AdaVD with SDXL v1, for long and semantically rich prompts that (do not) contain the \textit{``Snoopy''} concept while with the target concept \textit{``Snoopy''} to erase.
Although the prompts are more complex, they do not appear challenging for AdaVD to handle. 
AdaVD can still accurately identify and extract the relevant semantic components associated with the target concept, and can precisely erase these without affecting the background generation. 
We visualize the erased component for each generated image in the smaller images within each example block of Fig. \ref{fig:sdxl}, following the same approach as explained in the 2nd paragraph of Section \ref{sec:further}. 
These serve as supporting evidence, showing what semantic content has been removed by AdaVD.
For those prompts containing only the non-target concepts, AdaVD successfully retains nearly all the details of the non-target content, producing images that are virtually identical to those generated by the original SDXL v1.0.

\subsection{AdaVD on SDv3}
\begin{figure*}[!t]
\centering
\includegraphics[width=0.80\textwidth]{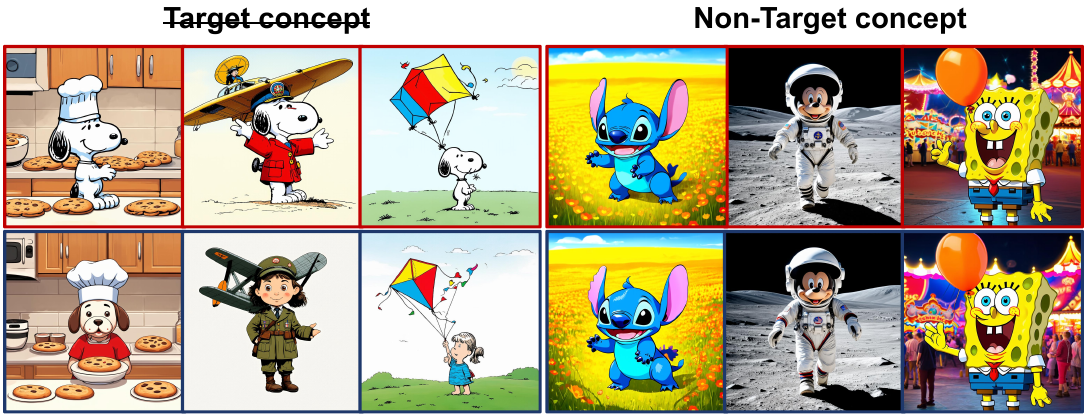}
\vspace{-2mm}
\caption{\textbf{Results of AdaVD on SDv3 for erasing \textit{``Snoopy"}}: The images with red and blue borders represent the before and after concept erasure, respectively. Our AdaVD effectively enables SDv3 to erase the target concept \textit{``Snoopy"} while preserving other semantic elements in the generated images. Moreover, AdaVD demonstrates outstanding prior preservation by ensuring that non-target concepts such as \textit{``Stitch"}, \textit{``Mickey"}, and \textit{``Spongebob"} remain highly similar to the generated images before concept erasure.}
\vspace{-3mm}
\label{fig:sd3}
\end{figure*}

A growing trend in text-to-image generative diffusion models is replacing U-Net with DiT as the noise predictor. Different from U-Net, DiT uses a transformer-based architecture, enhancing scalability in image generation. To validate the performance of our AdaVD in DiT-based diffusion models, we conduct experiments on SDv3. Different from SDv1.4 and SDXL, SDv3 uses the T5 text encoder \cite{raffel2020exploring}, alongside other encoders, to generate text embeddings for image generation. During the target embedding pre-processing phase, we handle text embeddings differently depending on the encoder: for embeddings from the CLIP text encoder, we replicate the last subject token, while for those from T5, we spread the mean embedding of all real word tokens. As shown in Fig. \ref{fig:sd3}, SDv3 successfully removes the target concept \textit{``Snoopy''} during the generation process while preserving the integrity of non-target concepts such as \textit{``Stitch''}, \textit{``Mickey''}, and \textit{``Spongebob''}. This highlights the strong prior preservation capability of AdaVD.

\begin{figure*}[!t]
\centering
\includegraphics[width=\textwidth]{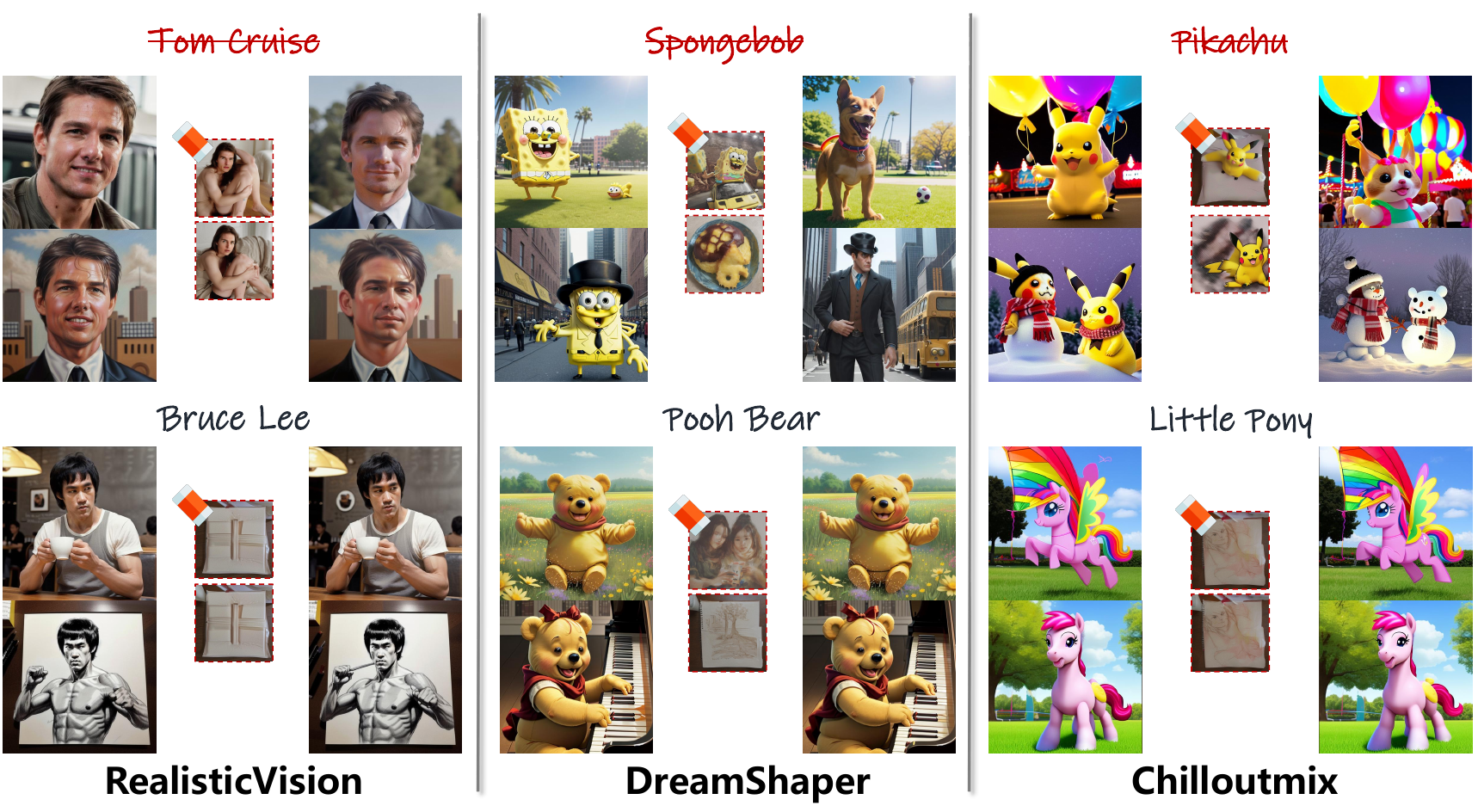}
\caption{\textbf{Results of AdaVD on other SD versions.} Our AdaVD illustrates a high performance of both erasure efficacy and the prior preservation across SD with difference versions and easing different concepts.}
\label{fig:other_version}
\end{figure*}

\subsection{AdaVD on Community SD Versions}

We also couple AdaVD with several community versions of SD, including RealisticVision \cite{RealisticVsion},  Dreamshaper \cite{DreamShaper}, and Chilloutmix \cite{Chilloutmix}, which are all fine-tuned based on SD v1.5. 
These versions target high-quality image generation with specific generation objectives. 
For example, RealisticVision specializes in generating lifelike images, while Dreamshaper excels in producing highly imaginative visuals. 
We experiment with removing the target concept \textit{``Tom Cruise''} from the text prompt corresponds to \textit{``Tom Cruise''} and \textit{``Bruce Lee''} for RealisticVision, removing \textit{``Spongebob''} from the text prompt corresponds to \textit{``Spongebob''} and \textit{``Pooh Bear''} for Dreamshaper, and removing  \textit{``Pikachu''} from the text prompt corresponds to \textit{``Pikachu''} and \textit{``Little Pony''} for Chilloutmix.

Fig. \ref{fig:other_version} presents the generated image examples.
The results show that AdaVD is capable of effectively erasing the target concept while preserving the integrity of the non-target content.
For all the experimented community versions, AdaVD can precisely locate the semantic space aligned with the target concept and isolate it with minimal disruption to the non-target semantics. 
Fig. \ref{fig:other_version} also visualizes the erased components as the smaller images within each example block, as in Fig.   \ref{fig:sdxl}.
Overall, the visualized erased components for prompts containing the target concepts show a high similarity to the target semantics. In contrast, for prompts corresponding to non-target concepts, the erased components lack meaningful semantic information.
These serve as additional evidence, showing the effectiveness of AdaVD.

\begin{figure*}[!t]
\centering
\includegraphics[width=0.9\textwidth]{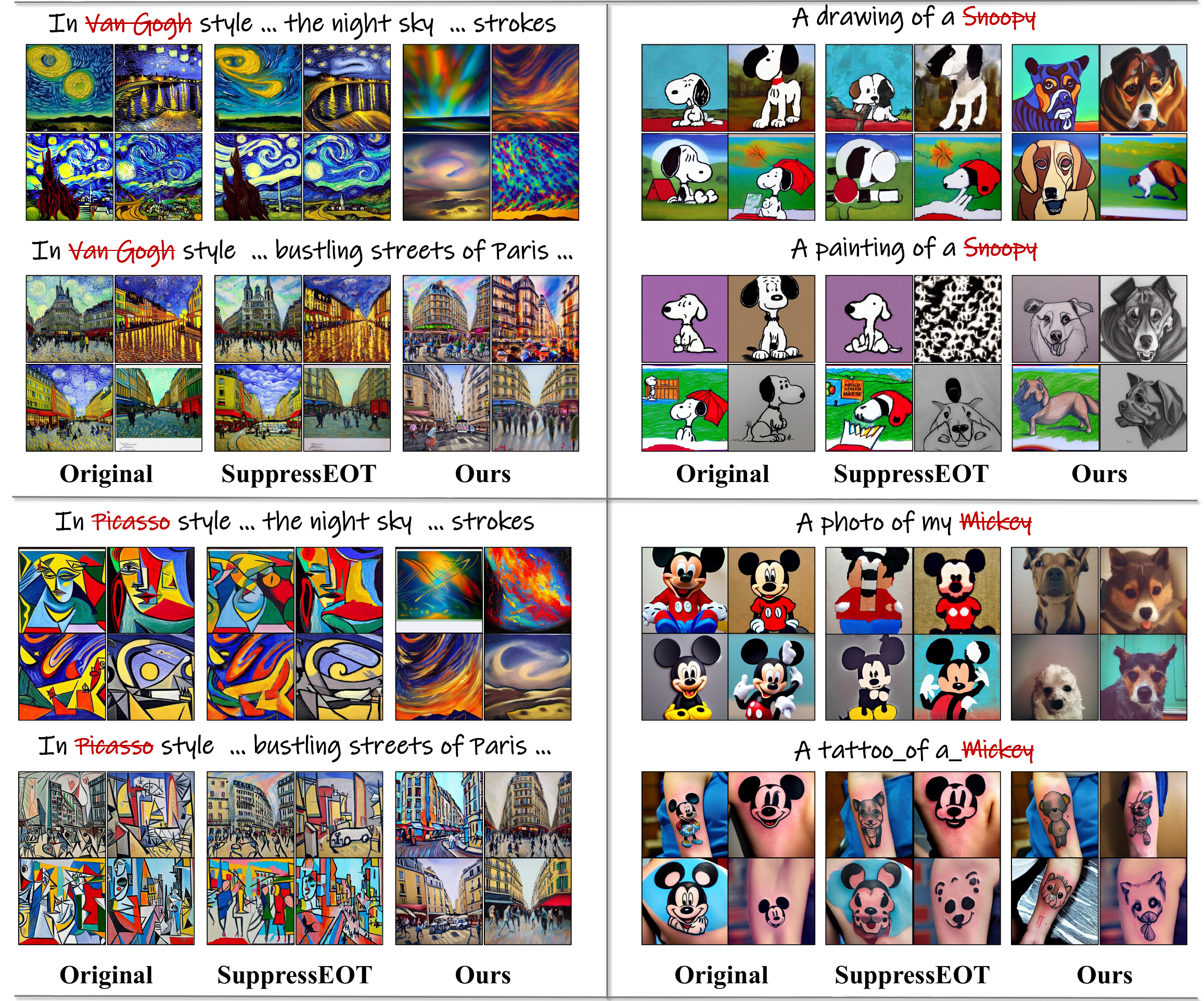}
\caption{\textbf{Qualitative comparison between SuppressEOT and  AdaVD.} We compare our AdaVD with SuppressEOT in single instance concept and art style erasure, demonstrating that AdaVD achieves more precise and effective erasure.}
\label{fig:suppresseot}
\end{figure*}

\section{Comparison with Additional Baselines}
\subsection{Comparison with SAFREE}
Orthogonal complement is widely used to decouple and separate out unwanted information. The art of using the orthogonal complement for concept erasure is on designing/deciding what space/direction to apply orthogonal complement, how to adjust removal strength, how to embed orthogonal complement in an algorithm to optimize its effect, etc. There is a concurrent work, SAFREE \cite{yoonsafree}, which also used orthogonal complement to facilitate concept erasure, but in completely different ways.
 Our AdaVD performs orthogonal complement in value spaces of attention layers within a diffusion model. Due to its effectiveness, there is no need for any complementary design, but a soft control of removal strength through a shift factor. Different from our AdaVD, SAFREE performs orthogonal complement over diffusion model input, i.e., text embedding space.
This approach necessitates complementary design elements, such as masking, another projection, and modifying detoxified embedding by Fourier transform. To control removal strength, it also uses a hard selection of whether to adopt the final detoxified embeddings.
We also conduct experiments to compare the performance of AdaVD and SAFREE in erasing art style concepts. As shown in Table \ref{tab:style_extend},  AdaVD achieves excellent prior preservation performance and second-best erasure efficacy, outperforming SAFREE, especially in prior preservation.

\subsection{Comparison with SuppressEOT}
\label{sec:gwyw}

In this additional experiment, we compare with a special concept erasure method SuppressEOT \cite{liget}, which requires the users to specify the positions of the erased concepts within the prompt.
Because of this user-involved setting, SuppressEOT is only applicable to specific prompts and is unable to achieve system-wide concept erasure.
Therefore, we only conduct a qualitative comparison of the erasure efficacy.  
Results are reported in Fig. \ref{fig:suppresseot}, where a comparison of art style erasure is shown on the left side, while the instance erasure results are displayed on the right.

It can be seen from Fig. \ref{fig:suppresseot} that AdaVD is precise and effective in erasing various concepts, achieving significantly higher erasure performance across a diverse range of use cases. 
Unfortunately, SuppressEOT consistently fails to remove completely the target concept from the generated images. 
For instance, when erasing \textit{``Mickey''} and \textit{``Snoopy''}, SuppressEOT is not even able to erase the general outlines of these specific instances. 
The unsatisfactory erasure performance of SuppressEOT likely stems from the fact that it was originally designed for image editing rather than concept erasure.
In image editing scenarios, preserving all the details of a prompt except for the target concept is important.
This is different from the requirement of concept erasure, where the prior preservation is needed only for the generation of non-target content. 
Such a difference in design requirement can inherently compromise the erasure efficacy of SuppressEOT.

\section{Additional Experiments and Analysis on Multi-Concept Erasure}
\label{appendix:multi-concept}

\begin{figure*}[!t]
\centering
\includegraphics[width=\textwidth]{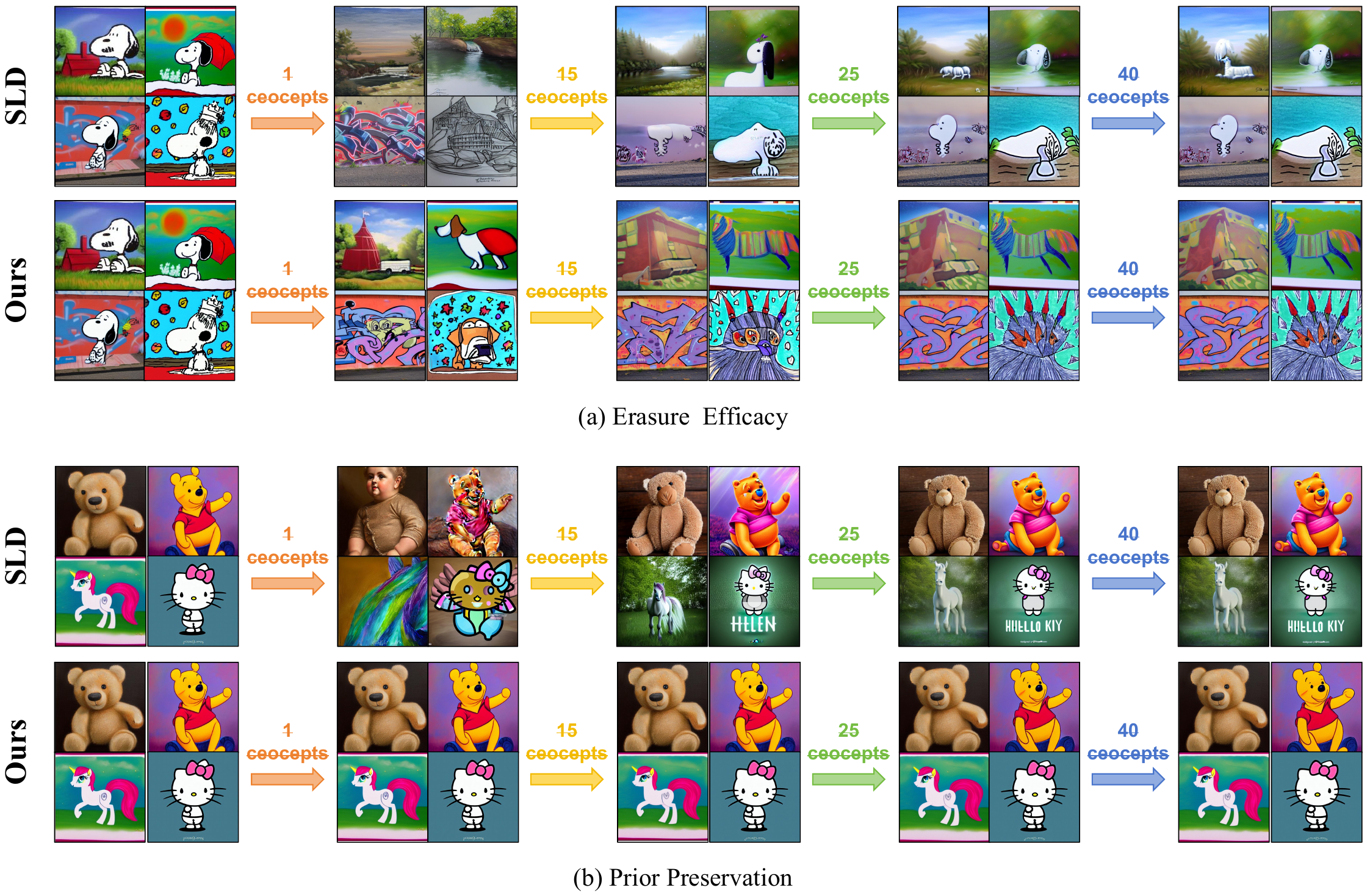}
\caption{\textbf{Examples of generated images for multi-concept erasure.} The illustrated examples show a consistently high performance of AdaVD in both erasure efficacy and prior preservation as the number of erased concepts increases, as compared to SLD.}
\label{fig:appendix-multi}
\end{figure*}
\begin{table*}[!h]
\centering

\begin{tabular}{c|>{\centering\arraybackslash}p{15cm}}
\toprule
\textbf{Number} & \textbf{Target Concepts}                                                                                                                                                                                                                                                                                                                                                                                                                                                                                                            \\ \midrule
1      & \textit{Snoopy}                                                                                                                                                                                                                                                                                                                                                                                                                                                                                                                     \\ \midrule
15     & \textit{Snoopy, Mickey, Crystal, Pikachu, Legislator, Bruce Lee, Marilyn Monroe, Tom Cruise, Anne Hathaway, Melania Trump, Van Gogh, Picasso, Rembrandt, Andy Warhol, Caravaggio}                                                                                                                                                                                                                                                                                                                                                   \\ \midrule
25     &\textit{ Snoopy, Mickey, Crystal, Pikachu, Legislator, Bruce Lee, Marilyn Monroe, Tom Cruise, Anne Hathaway, Melania Trump, Van Gogh, Picasso, Rembrandt, Andy Warhol, Caravaggio, Samoyed, Doraemon, Tom, Adam Driver, Adriana Lima, Amber Heard, Amy Adams, Andrew Garfield, Angelina Jolie, Anjelica Huston}                                                                                                                                                                                                                      \\ \midrule
40     & \textit{Snoopy, Mickey, Crystal, Pikachu, Legislator, Bruce Lee, Marilyn Monroe, Tom Cruise, Anne Hathaway, Melania Trump, Van Gogh, Picasso, Rembrandt, Andy Warhol, Caravaggio, Samoyed, Doraemon, Tom, Adam Driver, Adriana Lima, Amber Heard, Amy Adams, Andrew Garfield, Angelina Jolie, Anjelica Huston, Bradley Cooper, Bruce Willis, Bryan Cranston, Cameron Diaz, Channing Tatum, Charlie Sheen, Charlize Theron, Chris Evans, Chris Hemsworth, Chris Pine, Barack Obama, Beth Behrs, Bill Clinton, Bob Dylan, Bob Marley} \\ \bottomrule
\end{tabular}

\caption{\textbf{Number of concepts to be erased and their corresponding lists.} The number of concepts ranges from 1 to 40, demonstrating the efficacy of AdaVD in handling multi-concept erasure.}
\label{tab:concept}
\end{table*}

\subsection{On Erasing More Multi-concepts }

We conduct additional experiments,  investigating how our approach performs as the number of erased concepts increases, under a progressive setting.
We evaluate our AdaVD by first erasing one concept \textit{``Snoopy''} and gradually increasing the number of erased concepts to 15, 25, and 40. The details of the concepts to be erased for each case are listed in Table \ref{tab:concept}.
We work with the base T2I model SD v1.4 and compare it with the existing approach SLD.
The results are presented in Fig. \ref{fig:appendix-multi}, which extends Fig. \ref{fig:intro}.

It can be observed from the top erasure efficacy block of Fig. \ref{fig:appendix-multi}   that SLD gradually loses its precision when removing the target concepts. 
This is possible because SLD concatenates the target concepts into a prompt for guiding the generation process. 
When erasing too many concepts, the text encoder struggles to focus on each individual concept, resulting in diminished erasure efficacy. 
Additionally, some concepts may be truncated due to the token length limitation of the text encoder’s tokenizer. 
Differently, AdaVD achieves consistently high performance in multi-concept erasure. 
It constructs a value subspace based on the orthogonal complement of all the target concepts, which ensures that no information regarding any individual concept is lost. 

The bottom prior preservation block of Fig. \ref{fig:appendix-multi} shows that AdaVD is able to generate images nearly identical to the original ones, demonstrating a superior performance in prior preservation. 
But SLD struggles to preserve prior knowledge, for not only the more challenging case of removing a high number of concepts but also the simple case of removing one single concept.
It is worth noting that some slight change can be accumulated and amplified as the number of erased concepts increases, as shown in the hands and mouth of the generated image of \textit{``Pooh Bear''} by our AdaVD. 
Also, small pixel-level changes may grow into catastrophic forgetting with an increasing number of erased concepts due to error accumulation. 
Therefore it is important to use FID to evaluate the performance of prior preservation, as images that closely match the originals at pixel level should result in a low  FID score.

\begin{figure*}[!t]
\centering
\includegraphics[width=\textwidth]{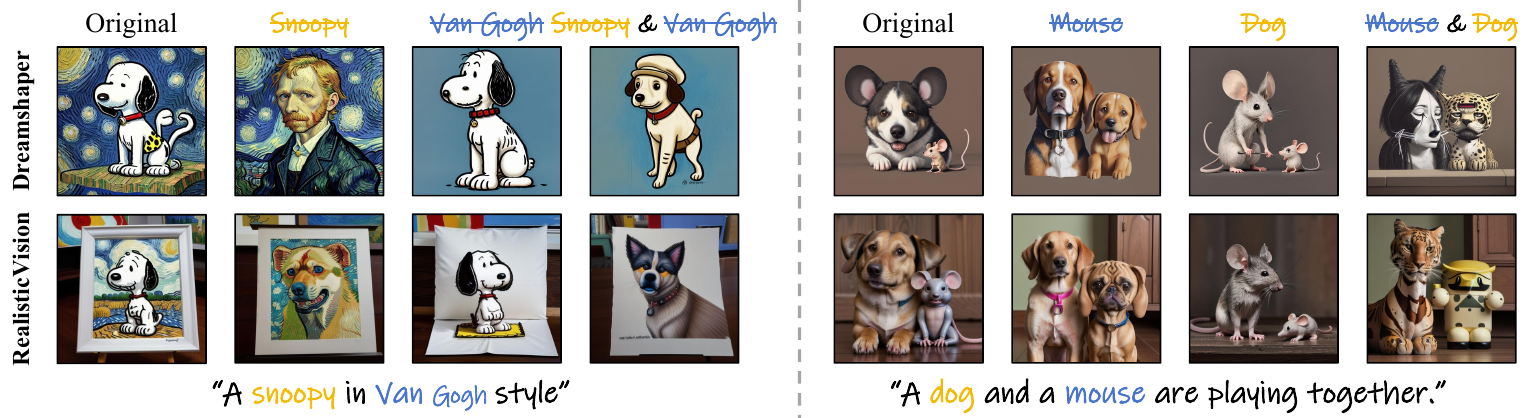}
\caption{\textbf{Results of AdaVD on multi-concept erasure across different SD versions.} We assess the performance of AdaVD on multi-concept erasure across various community versions of SD under diverse erasure scenarios, including cross-application erasure as outlined in SPM \cite{lyu2024one} and multi-instance erasure. These evaluations further highlight the robustness and effectiveness of AdaVD in addressing the challenges of the multi-concept erasure task.}
\label{fig:appendix-cross}
\end{figure*}
\begin{figure*}[!t]
\centering
\includegraphics[width=\textwidth]{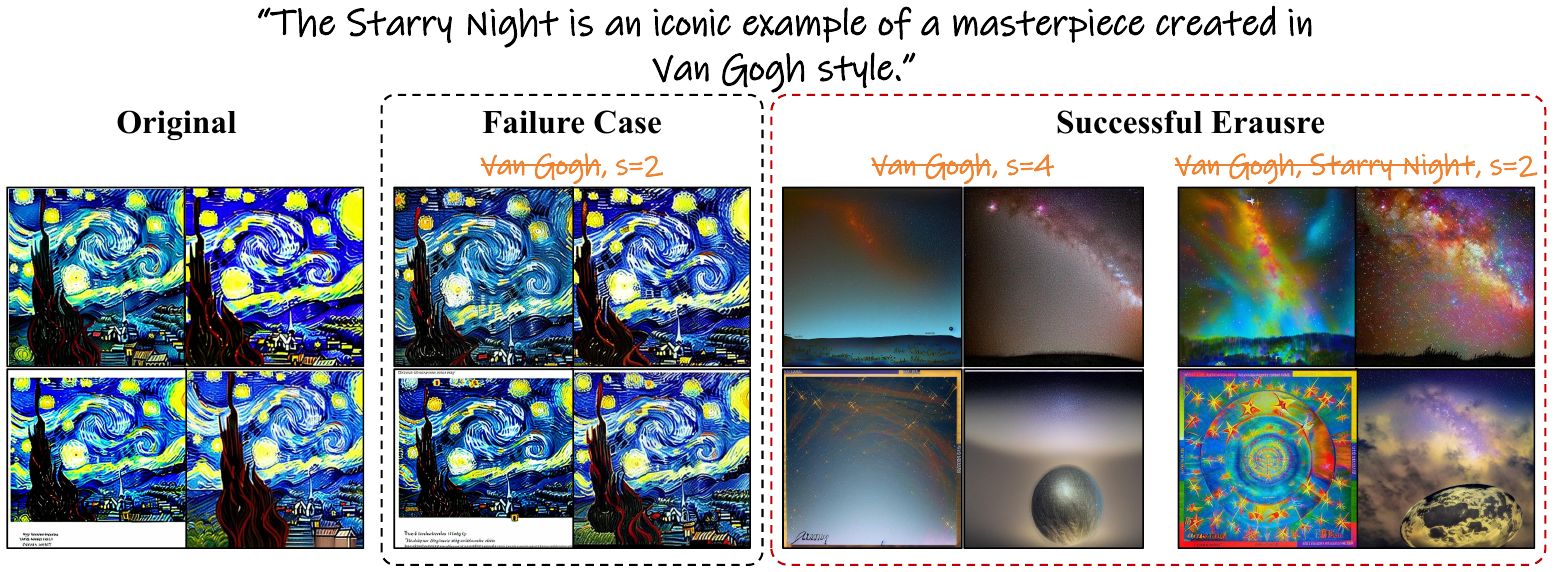}
\caption{\textbf{Failure case} when erasing \textit{``Van Gogh''} and its solution.}
\label{fig:fail}
\end{figure*}

\subsection{On Transferability to Other T2I Models}
In this additional experiment, we integrate AdaVD with two other T2I diffusion models, including DreamShaper \cite{DreamShaper} and RealisticVision \cite{RealisticVsion}, assessing its multi-concept erasure performance. 
Two multi-concept erasure scenarios are experimented with: one is cross-application erasure as described in SPM \cite{lyu2024one}, and the other is multi-instance erasure.
Results of the cross-application erasure are presented in the top half of Fig. \ref{fig:appendix-cross}, demonstrating the generated images after erasing \textit{``Snoopy''}, \textit{``Van Gogh''}, and the two concepts together. 
Results of the multi-instance erasure are shown at the bottom of Fig. \ref{fig:appendix-cross}, demonstrating the generated images after erasing \textit{``Mouse''}, \textit{``Dog''}, and both concepts. 
Overall, AdaVD achieves a high erasure precision. 
It can be seen from Fig. \ref{fig:appendix-cross} that, when aiming at a single concept erasure, other concepts specified in the prompt remain faithfully in the generated image; and when aiming at erasing multiple concepts, all the relevant visual content is also removed successfully.
This serves as evidence that AdaVD is capable of a robust and precise erasure. 

\section{Failure Case Study}
\label{sec:fail}
Despite its success, there exist concepts that AdaVD struggles to erase. We present a few failure cases in  Fig.~\ref{fig:fail}.
For instance, it is challenging for AdaVD   to erase \textit{``Van Gogh''} from a prompt like ``The Starry Night is an iconic example of a masterpiece created in Van Gogh style.'' 
The challenge is likely to stem from the presence of multiple tokens, \eg, \textit{``Starry Night''}, that is highly coupled with the target concept. 
In this case, a small value of the scaling hyper-parameter $s$ as used by the shift factor in Eq. (\ref{eq:shift_factor}) is insufficient to eliminate effectively the target semantics across all the relevant tokens. 
Nevertheless, this issue can be mitigated by doubling $s$ or incorporating additional related target concepts to erase, \eg, \textit{``Starry Night''}, evidenced by the right side of Fig.~\ref{fig:fail}.

\end{document}